\documentclass[10pt,letterpaper]{article}

\usepackage{amsmath,amssymb,mathtools}

\usepackage{changepage}

\usepackage[utf8x]{inputenc}

\usepackage{textcomp,marvosym}

\usepackage{cite}

\usepackage{nameref,hyperref}

\usepackage{microtype}
\DisableLigatures[f]{encoding = *, family = * }

\usepackage[table]{xcolor}

\usepackage{array}

\newcolumntype{+}{!{\vrule width 2pt}}

\newlength\savedwidth

\makeatletter
\renewcommand{\@biblabel}[1]{\quad#1.}
\makeatother

\usepackage{lastpage,fancyhdr,graphicx}
\usepackage{epstopdf}
\fancyheadoffset[L]{2.25in}
\fancyfootoffset[L]{2.25in}
\usepackage{booktabs,url,multirow}

\newtheorem{definition}{Definition}
\newtheorem{proposition}{Proposition}

\begin{document}
\vspace*{0.2in}

\begin{flushleft}
{\Large
\textbf\newline{False perfection in machine prediction: Detecting and assessing circularity problems in machine learning} 
}
\newline
\\
Michael Hagmann\textsuperscript{1\Yinyang*},
Stefan Riezler\textsuperscript{1,2\Yinyang*},
\\
\bigskip
\textbf{1} Computational Linguistics, Heidelberg University, 69120 Heidelberg, Germany
\\
\textbf{2} Interdisciplinary Center for Scientific Computing (IWR), Heidelberg University, 69120 Heidelberg, Germany
\bigskip

%
%
\Yinyang These authors contributed equally to this work.

* hagmann@cl.uni-heidelberg.de; riezler@cl.uni-heidelberg.de

\end{flushleft}
\section*{Abstract}
Machine learning algorithms train models from patterns of input data and target outputs, with the goal of predicting correct outputs for unseen test inputs. Here we demonstrate a problem of machine learning in vital application areas such as medical informatics or patent law that consists of the inclusion of measurements on which target outputs are deterministically defined in the representations of input data. This leads to perfect, but circular predictions based on a machine reconstruction of the known target definition, but fails on real-world data where the defining measurements may not or only incompletely be available. We present a circularity test that shows, for given datasets and black-box machine learning models, whether the target functional definition can be reconstructed and has been used in training. We argue that a transfer of research results to real-world applications requires to avoid circularity by separating measurements that define target outcomes from data representations in machine learning. 


\section*{Introduction}

Machine learning is a research field that has been explored for several decades, and recently has begun to affect many areas of modern life under the reinvigorated label of artificial intelligence. The goal of machine learning is to train models that can efficiently learn a wide class of input-output patterns to correctly determine outcomes on previously unseen input examples. The usual workflow of a machine learning research project consists of optimizing a machine learning model on given training data, tuning hyper-parameters on development data, and testing the final model using a standard automatic evaluation on benchmark test data. 
Due to the increasing availability of shared and curated data, this paradigm has significantly boosted research progress by fostering friendly competition between researchers and increasing comparability of results. The division of labor between the roles of data curator and machine learning practitioner also implies that the practitioner only needs to focus on finding a prediction model that improves the current state-of-the-art system on the shared data, without having to ask any questions about the origin, nature or compilation process of the data themselves. Furthermore, the emphasis on improving the state-of-the-art has led to a dominance of complex, non-intelligible neural network models in nearly all application areas.

In this article, we point out a problem of this practice that hampers the validity of machine learning research for a potentially wide range of applications.  We will illustrate this potential threat for two important areas of application, namely the vital application area of (early) disease diagnosis in medicine, and the economically important application of patent prior-art search. Both fields are examples of empirical sciences that define the objects of their research by rigid measurement procedures and empirically testable criteria. For instance, sepsis, a disease that causes about $20$\% of all global deaths \cite{RuddETAL:20}, is operationalized in the currently used Sepsis-3 definition as a change in total SOFA (sequential organ failure assessment) score of at least 2 points consequent to an infection. The SOFA scoring system \cite{VincentETAL:96,SingerETAL:16,SeymourETAL:16} is defined for 6 organ systems whose scores are defined by thresholds on measurable physiological quantities like heart rate, creatinin, bilirubine, urine output etc.  As shown in a recent overview \cite{MoorETAL:20} that examined 22 studies on machine learning approaches for (early) prediction of sepsis, with the exception of one, all studies define ground-truth sepsis labels using the deterministic rules of the consensus definition of Sepsis-3  \cite{SingerETAL:16,SeymourETAL:16} or Sepsis-2
\cite{LevyETAL:03,DellingerETAL:13}, and use datasets such as MIMIC-II or MIMIC-III \cite{JohnsonETAL:16} or the Emory University Hospital data \cite{ReynaETAL:19} for training and testing. However, these datasets 
include time series of all clinical measurements that are used in the definition of sepsis according to Sepsis-3 or Sepsis-2. We argue that including measurements  which are  known to be deterministically related to the target label in the feature representations of the input data leads to a circularity in prediction where the machine learning model learns nothing else but to reconstruct the known functional definition (in this case) of sepsis. However, the true goal of applying machine learning to (early) prediction of sepsis is to construct a system that captures patterns beyond a straightforward application of what is already known, for example, for unseen inputs that exhibit uncommon patterns or where the measurements applied in the sepsis definitions are incomplete.

A similar problem is evidenced in the area of patent prior-art search. This task is economically extremely relevant, and describes the need to cite all previous patents that are relevant to the claim of the originality of a new patent in filing a patent application.  The machine learning use case is to aid the patent inventor or patent examiner by automatic search for prior-art with respect to a query patent over a search repository of (possibly foreign-language) patents documents. An established technique to create large datasets for research in patent retrieval is to determine relevance ranks from different types of patent citations \cite{GrafAzzopardi:08,PiroiTait:10,GuoGomes:09}: Highest relevant patents are those in the same patent family, indicating the same invention disclosed by common inventors and patented in more than one country (relevance level $r=3$); very relevant patents are the ones cited in search reports by patent examiners ($r=2$); lowest relevant patents are  citations added by patent applicants ($r=1$). In such scenarios a problem of {circular features} arises if the criteria that are used to define the gold standard relevance labels are incorporated as features in the data of machine learning models. This happened, for example, in the CLEF-IP 2010 benchmark competition \cite{PiroiTait:10} where applicant citations extracted from the query document were added to the list of retrieval results in the approach of \cite{MagdyJones:10}. Because of overlapping citations in the examiner citations and applicant citations, this procedure has been criticized as raising a concern about the validity of the evaluation of the task \cite{MahdabiCrestani:14}.

\subsection*{A circularity test for machine learning data and models}

In the following, we present a simple statistical method to assess circularity that is based on fitting \emph{generalized additive models (GAMs)} \cite{HastieTibshirani:90,Wood:17} to data $D = \{(\mathbf{x}^n,{y}^n)\}_{n=1}^N$ consisting of $N$ pairs of input features $\mathbf{x}^n$ and output labels $y^n$, where the latter are either gold-standard labels from machine learning data, or predicted labels from machine learning models. GAMs are an additive combination of functions $f_k(x_k)$ called \emph{feature shapes}, that decompose a complex function into one-dimensional components $f_k$ for each feature $x_k$ (or pairs of features $(x_i,x_j)$) in a $p$-dimensional input representation $\mathbf{x}=(x_1, x_2, \ldots, x_p)$. GAMs have a high expressive power since the feature shapes are nonlinear functions, however, the model is intelligible since the contribution of each feature $x_k$ to the prediction can be interpreted by visualizing feature shapes via plotting $f_k(x_k)$ against $x_k$.
Given machine learning training data, the goal of the GAM-based assessment is twofold. Firstly, it consists in assessing whether a deterministic rule that defines the target (and especially the features associated with it) can be identified and extracted from the data set in order to prevent the machine learning system to learn what is already known. In this case we assume that the functional definition of the target is unknown (to the machine learning practitioner), and we want to learn whether the training data include the necessary information to extract a deterministic rule.  Secondly, this method is used to assess the predictions of a black-box machine learning model with the goal of investigating the patterns a system has picked up during training. In that case we assume that we have access to model predictions for test data, and we treat the machine learning system as an unknown function we would like to study. This application is not only of interest to the machine learning practitioner, but also (and even more so) to the potential model operator.  For example, a model operator who knows the deterministic function used to generated the ground truth labels may want to assess if the system has learned to make correct predictions beyond solely applying the known target definition.  

Irrespective of the analytic goal, our method is based on two measures. The first measure allows assessing the goodness-of-fit of a GAM to its training data. For this purpose we use the likelihood-based measure of \emph{scaled deviance}, denoted by $D^2$. This measure ranges between $0$ and $1$, where a value of $1$ indicates that our model can replicate the data perfectly, while a value of $0$ corresponds to a model that predicts a constant mean target value for all inputs.  We then define a set of candidate models --- if feasible, we suggest to consider all possible models --- which we fit to the data and rank the fitted models according to deviance and model complexity. We are searching for a model with low complexity that still achieves a deviance value close to  $1$.  A second criterion is called \emph{nullification} of a feature, expressed by a constant zero feature shape in a GAM. By exploiting the consistency property of the maximum likelihood estimator used to fit GAMs, we can use this second criterion to distinguish input features that are deterministically related to the labels in the data from possible additional features in the input data that are irrelevant to the label-defining function. Based on the measures of deviance and nullification, we will define a circularity test that allows identifying circular features as features identifying a deterministic data generation model. More information on the mathematical details will given in the Methods section below. 

In the Experiments section, we will demonstrate our argumentation by applying the circularity test to the building blocks of sepsis prediction, namely prediction of SOFA scores for liver and kidney. We will show that the target functional definition of SOFA scores can exactly be reproduced, even from predictions of a black-box machine learning model, if the respective clinical measurements have been included in the data. Moreover, the contribution of all other features consisting of clinical and laboratory measurements is completely nullified. This shows firstly that machine learning models trained on data including the defining measurements for SOFA scores will yield nearly perfect predictions on input data that include the defining measurements, but cannot be transferred to unseen data where the defining features are not or only incompletely available. Secondly, a circular learning setup that nullifies the contribution of all features except the ones defining the target, dashes the hope to detect features that could shed a new light on sepsis prediction. 

A similar result can be exemplified by applying a circularity test to data and machine learning models for patent prior-art search: If the data contain citation information, and if this information is used to define gold-standard relevance labels, then this defining feature can be identified as the sole feature that is responsible for nearly perfect predictions. In addition, the contribution of otherwise strong features such as similarity functions on patent query and candidate prior-art patents are nullified. Such a model can only be applied if citation information is included in the input representations, and fails completely if this information is not or only sparsely available. 

\subsection*{Related Work}

Our circularity test is based on GAMs as a tool to identify the use of circular features in training of black-box models such as neural networks. While GAMs have been known as intelligible white-box models for machine learning \cite{LouETAL:12} and as tools for analyzing black-box models \cite{TanETAL:18}, they have not been used to identify circular features before. 

Our work is related to, but different from recent work on interpretability or explainability in machine learning \cite{Doshi-VelezKim:17,Miller:17,RibeiroETAL:16}. Such techniques have been applied in natural language processing or image processing to identify features that rely on simple, spurious patterns (see, for example, \cite{LapuschkinETAL:19,SchlegelETAL:20}). However, these so-called bias features are essential to the overall task, so they cannot simply be ignored, but must be incorporated into robust models that minimize their influence in sophisticated ways (see, for example, \cite{KimETAL:19,ClarkETAL:19}). 

Another related area is that of so-called illegitimate features \cite{KaufmanETAL:11}. For example, based on suspicion that information was inadvertently leaked by compiling data from different medical institutions, and collecting positive and negative labels from particular institutions with consecutive patient IDs, \cite{RossetETAL:09} could identify patient IDs as features that are highly correlated with the target label.
In difference to illegitimate features, circular features are characterized by a legitimate deterministic relation to the target labels. On the other hand, if the target functional definition is known, incorporating it as information source in training and inference of machine learning models defeats the purpose of machine learning --- transferring patterns learned on training data to unseen test data, and detecting new enlightening patterns in the process.

\section*{Methods}

A prerequisite for our circularity test is the availability of an expressive and yet interpretable model that can be fitted to data $D = \{(\mathbf{x}^n,{y}^n)\}_{n=1}^N$ resulting from a data annotation or model prediction process.  
As a model we adopt the highly expressive and yet interpretable class of \emph{Generalized Additive Models} (GAMs) that originated in the area of biostatistics \cite{HastieTibshirani:90} to circumvent the restriction of strictly linear features in generalized linear regression models.  The key idea of GAMs is decomposing a multivariate function into a sum of functions with lower dimensional inputs, called \emph{feature shapes}, which are learned from the data. The one-dimensional feature shapes $f_k(x_k)$ for each feature $x_k$ (or pairs of features $(x_i,x_j)$) in $\mathbf{x}=(x_1, x_2, \ldots, x_p)$ are additively combined and can be nonlinear functions themselves. 
The general form of a GAM assumes $Y$ to be a random variable from the exponential family, and $g(\cdot)$ to be a nonlinear link function: 
\begin{align}
g(\mathbb{E}[Y|\mathbf{x}]) = \sum_{k=1}^p f_{k}(x_{k}) + \sum_{i \ne j} f_{ij}(x_{i},x_{j}).
\end{align}
An additive Gaussian is obtained by using the identity link function $g(x)=x$, and specifying the distribution of $Y^n$ to be of the Gaussian subclass of the exponential family:
\begin{align}
Y^n = \sum_{k=1}^p f_{k}(x_{k}^n) + \sum_{i \ne j} f_{ij}(x_{i}^n,x_{j}^n) + \epsilon^n, \textrm{ where } \epsilon^n \sim \mathcal{N}(0,\sigma^2) \textrm{ for } n = 1, \ldots,N.
\end{align}

In the following examples, feature shapes are modeled by regression \emph{spline} functions. For mathematical background on modeling with splines and on estimation of GAMs we refer the reader to \cite{HastieTibshirani:90,Wood:17}. 

The first part of our statistical test for circularity based on GAMs is a measure for the fit of the model to the data $D = \{(\mathbf{x}^n,{y}^n)\}_{n=1}^N$. We will use the likelihood-based criterion of \emph{scaled deviance} of a model for this purpose. \cite{McCullaghNelder:89} define it as a metric proportional to the difference between the log-likelihood $\ell(\mu)$ of a model $\mu$ to the log-likelihood $\ell^\ast$ of the saturated model, i.e., to the model in the distributional family that achieves the highest possible likelihood value given the data : 
\begin{align}
    D^\ast_\mu = 2(\ell^\ast - \ell(\mu)).
\end{align}
The saturated model corresponds to an exact fit by setting the fitted values equal to the observed data. The distribution of a single observation of the additive Gaussian model with known variance $\sigma^2$ is
\begin{align}
    p(y^n|\mu^n,\sigma^2) = \frac{1}{\sqrt{2\pi\sigma^2}} \exp(-\frac{(y^n-\mu^n)^2}{2\sigma^2}),
\end{align}
with log-likelihood 
\begin{align}
    \ell(\mu^n) = -\frac{1}{2} \log (2\pi\sigma^2) - \frac{(y^n-\mu^n)^2}{2\sigma^2}.
\end{align}
Setting $\mu^n = y^n$ in the saturated model yields $\ell^\ast = -\frac{1}{2} \log (2\pi\sigma^2)$ so that 
\begin{align}
     D^\ast_\mu = 2(\ell^\ast - \ell(\mu)) = \frac{(y^n-\mu^n)^2}{\sigma^2}.
\end{align}
Apart from the scaling factor $\sigma^2$, this metric is identical to the residual sum of squares $R^2$, which is a standard measure of model fit in statistics.  Following \cite{HastieTibshirani:86}, we use the \emph{percentage of deviance explained} to make the metric more interpretable, and denote it by $D^2$ in analogy to $R^2$:
\begin{align}
    D^2(\mu) = 1 - \frac{D^\ast_\mu}{D_{\mu_0}}, 
\end{align}
where $D_{\mu_0}$ is the deviance for the model $\mu_o$ that uses just a constant intercept term (without any predictor variables) for all response variables, yielding $D^2(\cdot) \in [0,1]$. 

The intended usage of the $D^2$ metric in a circularity test on a given dataset $D = \{(\mathbf{x}^n,{y}^n)\}_{n=1}^N$ is to train a set of GAMs, one for each member of the powerset of features, and to find the model with maximal $D^2$ and smallest degrees of freedom.
Degrees of freedom of a model are calculated by the number of tuneable parameters. For example, a GAM for $n=1, \ldots,N$ data points, modeling feature shapes for each of $k=1, \ldots, p$ input features with cubic splines of $d_k$ parameters for each feature, together with a smoothness penalty for each of feature, adds up to $(N \times \sum_{k=1}^p d_k) + p$ degrees of freedom.

In addition to this, we employ a second check to differentiate input features that are deterministically related to the labels in the data from possible additional features in the input data that are irrelevant to the label-defining function. This check is called \emph{nullification} and is based on the consistency of maximum likelihood estimators for GAMs. The identifiablity of GAMs trained on data that have been produced by  a deterministic data labeling function allows identifying features $x_k$ that determine the data generation process by their non-zero feature shape $f(x_k)$ approximating the deterministic labeling function, while any additional features $x_j, j \neq k$ will have a feature shape of a constant zero function. A more formal account on nullification is given in Section A.1 of the Appendix.

We define a circularity test that proceeds by searching for the model with highest deviance and lowest degrees of freedom over the powerset of features, and by confirming that all other features except the ones found in the first step are nullified.

\begin{definition}[Circularity Test]
Given a dataset of feature-label relations $D = \{(\mathbf{x}^n,{y}^n)\}_{n=1}^N$ where $\mathbf{x}^n = (x_1, x_2, \ldots, x_p)$ is a $p$-dimensional feature vector, let $C \subseteq \mathcal{P}(\{1,\ldots, p\})$ indicate the set of candidate circular features in dataset $D$, and let $\mathcal{M} \coloneqq \{ \mu_c \colon c \in C\}$ be the set of models obtained by fitting a GAM based on feature set $c$ to the data $D$. A set of circular features  $c^*$ is detected by applying the following two-step test:
\begin{enumerate}
    \item $c^* = \operatorname*{argmax}_{c \in C} D^2(\mu_c)$ where $D^2(\mu_{c^*})$ is close to $1$, and in case the maximizer is not unique, the maximizer is chosen whose associated GAM $\mu_{c^*}$ has the smallest degrees of freedom.
    \item The feature shapes of any other features added to the GAM $\mu_{c^*}$ are nullified in the model $\mu_{\{1,\ldots, p\}}$ that is based on the full feature set.
\end{enumerate}
\end{definition}
Note that identifiability and thus consistency of maximum likelihood estimators is an essential property of spline-based GAMs as described in \cite{HastieTibshirani:90,Wood:17,Heckman:86}. This is not necessarily true of neural network-based NAMs \cite{AgarwalETAL:20}, for which identifiability or consistency has not been shown. Furthermore, note that the circularity test defined above is not restricted to single features, but it allows assessing the circularity of feature sets by fitting and testing the deviance and nullification of multivariate GAMs. 

\section*{Experiments}

\subsection*{Circularity in machine learning data}

In the following examples, we assume that we have access to a dataset $D = \{(\mathbf{x}^n,{y}^n)\}_{n=1}^N$ of input features $\mathbf{x} = (x_1, x_2, \ldots, x_p)$ and gold standard labels $y$. We assume that we do not know if a deterministic rule system has been used to assign gold standard labels, and what features are relevant for the deterministic labeling. Since including such features in a machine learning dataset and consequently, in a machine learning model, would effectively circumvent a meaningful prediction, our goal is to detect such features in order to prevent their usage as circular features.  We will show that the circularity test stated above allows us to detect such features if they exist in the dataset.

\subsubsection*{Circularity in patent prior art data.}

Let us consider as a first example for circularity the case of data annotation in cross-lingual patent retrieval. This task is a subclass of cross-lingual information retrieval and economically extremely relevant. If a company wants to file a patent application, it is important that the new patent cites all previous patents that are relevant to the claim of its originality. The task of identifying relevant patents is called ``patent prior art search''. In practice, the patent applicant adds all citations that are relevant, to the best of his knowledge, and then this list is refined by patent examiners specifically trained on certain areas of technology. The machine learning task is to aid the patent inventor or patent examiner by automatic prior art search for a set of patent queries over a search repository of foreign-language patents documents.

\begin{table}[t!]
\caption{Definition of relevance ranks based on patent citation information. Note that the definition is exclusive so that only one condition applies at a time.}
\label{tab:definition_relevance}
\begin{center}
\begin{tabular}{lc}
  \toprule
  Condition &  Relevance Score \\
  \midrule
  no citation &  $0$\\
  inventor citation &  $1$\\
  examiner citation  &  $2$\\
  family patent  & $3$\\
  \bottomrule
  \end{tabular}
\end{center}
\end{table}

Relevance ranks for training and testing in information retrieval can be obtained from human annotators who assign relevance ranks to query-document pairs \cite{ChapelleChang:11,QinETAL:10}, or by inferring relevance ranks from user clicks \cite{Joachims:02,AgrawalETAL:09}. An automatic procedure called pooling for assigning relevance ranks has been extablished in the Text REtrieval Conferences (TREC) \cite{VoorheesHarman:05}: Here the top $k$ documents returned by a number of different retrieval systems are assigned as relevant.
For patent prior art search, an established technique to create large datasets is to determine relevance ranks from different types of patent citations \cite{GrafAzzopardi:08,PiroiTait:10,GuoGomes:09}: The most relevant patents are those in the same patent family, indicating the same invention disclosed by common inventors and patented in more than one country (relevance level $r=3$); very relevant patents are the ones cited in search reports by patent examiners ($r=2$); lowest relevant patents are  citations added by patent applicants ($r=1$).  These four exclusive conditions build a rule system shown in Table \ref{tab:definition_relevance} that allows deterministically assigning relevance ranks to all query-document pairs. 

\begin{table}[t!]
\caption{Feature set for cross-lingual patent retrieval.}
\label{tab:features_clir}
\begin{center}
\begin{tabular}{lll}
\toprule            
Feature & Meaning & Range \\
\midrule
        \textbf{\texttt{neural}} & similarity score learned by neural network  & $\mathbb{R}$\\
        \textbf{\texttt{tf-Idf}} & cosine similarity of tf-Idf scores & $\mathbb{R}$\\
        \textbf{\texttt{inventor}} & indicator for inventor citation & $\{0,1\}$\\
        \textbf{\texttt{examiner}} & indicator for examiner citation & $\{0,1\}$\\
        \textbf{\texttt{family}} & indicator for family patent  & $\{0,1\}$\\
\bottomrule 
\end{tabular}
\end{center}
\end{table}

The data used in our experiment are a subset of the dataset used by \cite{KuwaETAL:20} for cross-lingual patent retrieval from Japanese to English. They consist of $425{,}065$ observations, including $2{,}000$ patent queries (each with around $250$ relevant documents at various levels of relevance) and $200$ sampled irrelevant documents per query. 
For our experiment, the data are split into a training set of $1{,}500$ queries (with $318{,}375$ observations of query-document pairs), and a test set of $500$ queries (with $106{,}690$ observations of query-document pairs). The dataset also contains two scores that measure the similarity between the query and a search document.  One is the cosine-similarity of the tf-Idf scores \cite{Jones:1972} from the Google-translated query and the search document, and the other is a learned similarity metric derived by \cite{KuwaETAL:20} by training a deep neural network on query and search document text and category data (henceforth called neural similarity score). The full list of input features of the dataset is shown in Table \ref{tab:features_clir}. Furthermore, the dataset contains a relevance score for each query-document pair which are the target labels for the prediction task. We know that the gold standard relevance ranks for the data used by \cite{KuwaETAL:20} have been produced following the deterministic rule shown in Table \ref{tab:definition_relevance}.  

Let us assume a research team wants to use this dataset to train a cross-lingual patent retrieval system without, however, knowing how the gold standard relevance rankings were defined. The goal of the research team is to find out if a deterministic procedure has been used to assign relevance ranks, and to reconstruct the deterministic rule system in an intelligible way, in order to avoid including features on which target labels are deterministically defined into their model. For this purpose, our research team applies the circularity test to the powerset of features that can be constructed from Table \ref{tab:features_clir}.

\begin{table}[t!]
	\caption{Top five models visited during circularity search for IR training data.}
	\label{tab:top5_circ_ir}	
	\begin{center}
		\begin{tabular}{cccc}
			\toprule
			Rank &  Included Features  & $D^2$ & Complexity \\
			\midrule
			1 & \{inventor, examiner, family\} & $100\%$ & $5$ \\
			2 & \{inventor, examiner, family,  neural\}& $100\%$ &$6.33$ \\
			3 & \{inventor, examiner, family,  tf-Idf\} & $100\%$ & $7.95$ \\
			4 & \{inventor, examiner, family, neural, tf-Idf\} & $100\%$ & $11.1$ \\
			5 & \{examiner, family, neural,  tf-Idf\} & $95\%$ & $22$ \\
			\bottomrule
		\end{tabular}
	\end{center}
\end{table}

\begin{figure}[t!]
	\centering
	\includegraphics[width=0.8\textwidth]{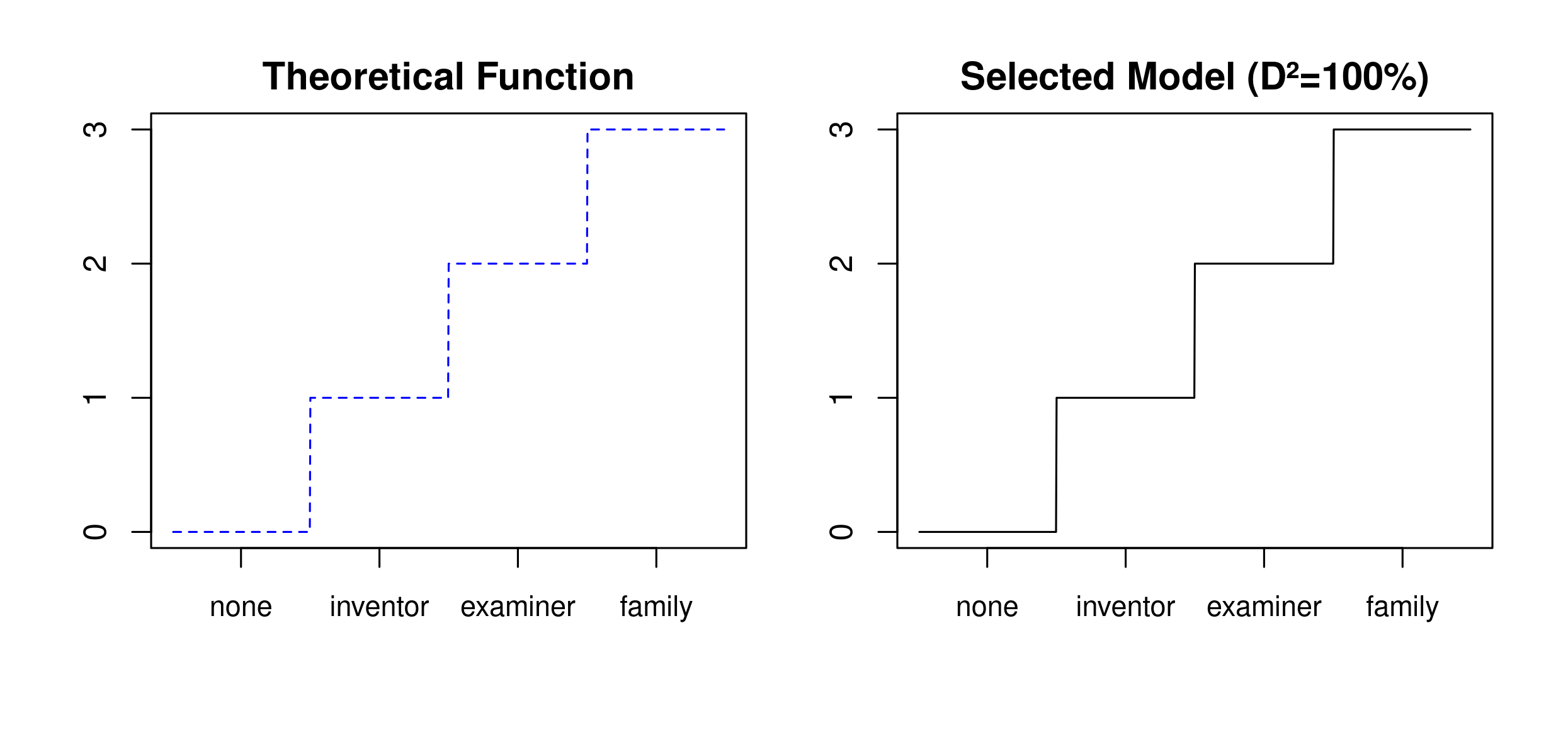}
	\caption{Feature shape of citation features reconstructing target labeling function.}
	\label{fig:clir_reconstructed_function} 
\end{figure}

Table \ref{tab:top5_circ_ir} shows the top five models trained during the search procedure. All models that include the citation features inventor, examiner, and family, perfectly reproduce the training data, as shown by values of $D^2=100\%$. The model consisting of only these three features, excluding tf-Idf or neural, is the least complex one. Furthermore, as shown in Fig \ref{fig:clir_reconstructed_function}, the feature shapes of these three features show that they perfectly reconstruct the target function.
 This is a first indicator that the set of citation features has been used to deterministically define the target labels, thus revealing them as potentially circular. 
 
\begin{figure}[t!]
	\centering
	\includegraphics[width=0.8\textwidth]{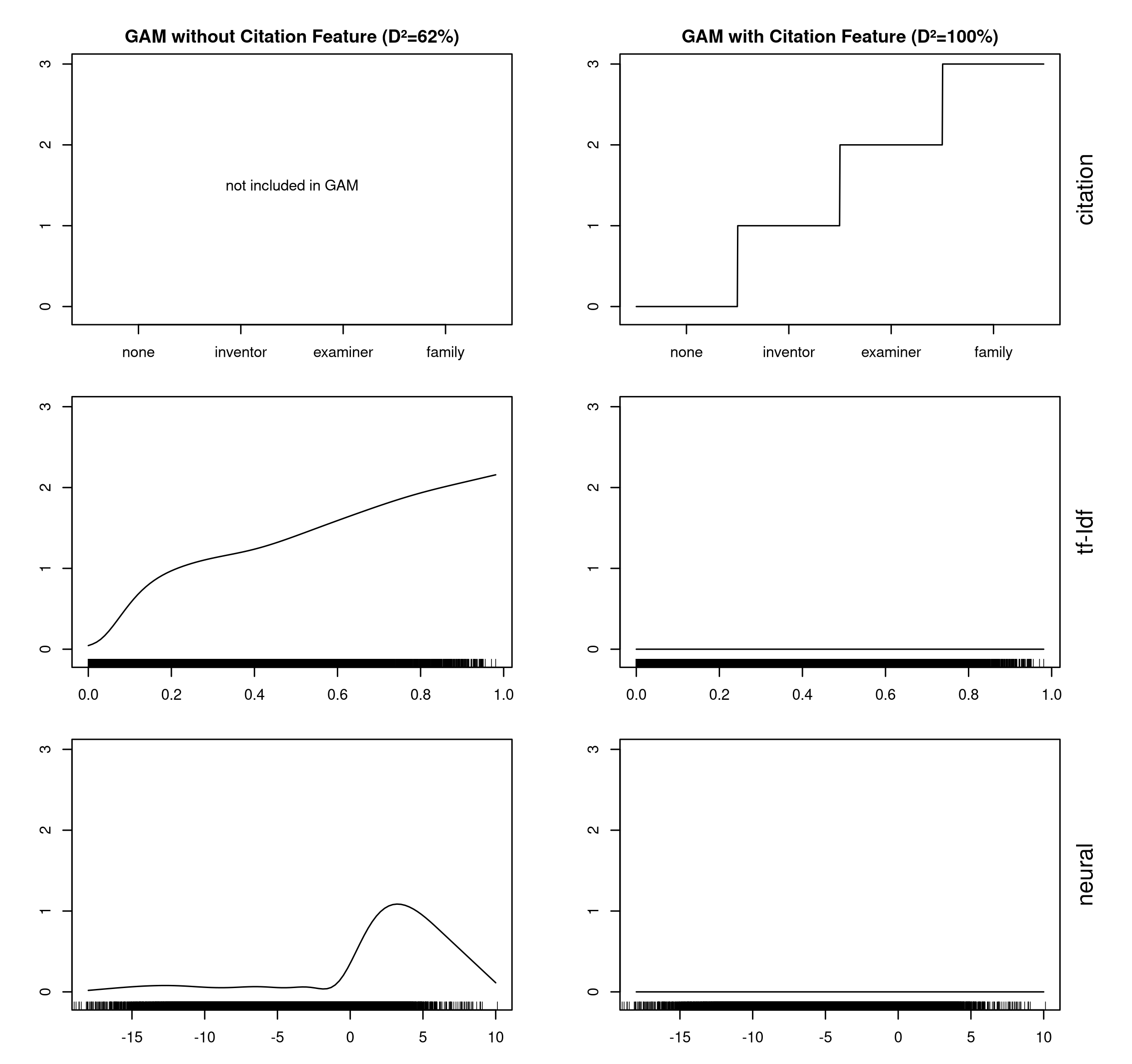}
	\caption{Feature shapes of GAM trained with (right column) and without access to citation information (left column), showing nullification of non-circular features in the presence of circular features.}
	\label{fig:clir_nullification} 
\end{figure}

Further circularity evidence is obtained by comparing the feature shapes of a model trained on all features to the feature shapes of one where the circularity candidates, i.e., the citation features, were omitted. 
The left column of Fig \ref{fig:clir_nullification} shows the feature shapes of the tf-Idf and neural features in a model without citation features. The strong contribution of these features to the prediction of relevance scores is visible with a $D^2$ value of $62\%$. For example, the plot on the middle left shows that relevance score is a nearly linear function of tf-Idf score. The top right of Fig \ref{fig:clir_nullification} shows the feature shape of the citation features for a model that includes all features. Like any model that includes citation features, this model has a $D^2$ value of $100\%$, and it allows us to exactly reconstruct the theoretical step function of relevance scores. However, as seen in the middle and bottom right of Fig \ref{fig:clir_nullification}, the contribution of the tf-Idf and neural feature in the model that combines all features is completely nullified. We note that this nullification in Fig \ref{fig:clir_nullification} is perfect in that the feature shapes of the nullified features are constant zero lines. This confirms our analysis of the citation features being circular in the investigated patent retrieval dataset. 

\subsubsection*{Circularity in medical data.} 

Another frequent case of circularity in data annotation is the measurement-based determination of gold standard labels in medical data science. A typical task in medical data science is the construction of machine learning based (early) disease diagnosis systems. Let us consider the case of sepsis which is a prevalent (especially among intensive care  patients) and lethal disease \cite{RuddETAL:20} whose early stages are hard to diagnose. Early diagnosis, however, is crucial to start an effective treatment. Since the introduction of the Sepsis-3 definition \cite{SingerETAL:16,SeymourETAL:16}, the Sequential Organ Failure Assessment (SOFA) score \cite{VincentETAL:96} has played a crucial role in sepsis diagnosis. Together with a suspicion of infection, a defining property of sepsis according to the Sepsis-3 definition is a change in total SOFA score $\geq 2$ points consequent to an infection, for SOFA scores defined for six organ systems. The SOFA scores are based on thresholds of clinical measurements. For example, the SOFA scores for the liver and the kidney are based on thresholding measurements of biochemical processes occurring in the respective organ systems.

The data set for our experiments were collected from $620$ intensive care patients from the surgical intensive care unit of the University Medical Centre Mannheim, Germany (see \cite{SchamoniETAL:19}) for a detailed description). Out of the $45$ features used in \cite{SchamoniETAL:19}, we consider the clinical measurements of bilirubin (bili), aspartate aminotransferase (asat), quick-inr (quinr), alanin aminotransferase (alat), and cardiac output (hzv) as possible features to describe the liver SOFA score. These features were selected based on the magnitude of their bivariate correlation with the liver SOFA score.  As shown in Table \ref{tab:definition_lsofa}, the deterministic rule to define the SOFA score for the liver is based solely on intervals of bilirubin values. 
 
 \begin{table}[t!]
 \caption{Definition of liver SOFA score based on bilirubin levels.}
	\label{tab:definition_lsofa}
	\begin{center}
		\begin{tabular}{cc}
  			\toprule
  			Condition & Liver SOFA Score \\
  			\midrule
  			$0 < $ bilirubin $\leq 1.2$ &  $0$\\
  			$1.2 < $ bilirubin $\leq 1.9$ &   $1$\\
  			$1.9 < $ bilirubin $\leq 5.9$ &  $2$\\
  			$5.9< $ bilirubin $\leq 11.9$  & $3$\\
  			bilirubin $> 11.9$  & $4$\\
  			\bottomrule
  		\end{tabular}
	\end{center}
\end{table}

\begin{figure}[t!]
	\centering
  		\includegraphics[width=0.5\textwidth]{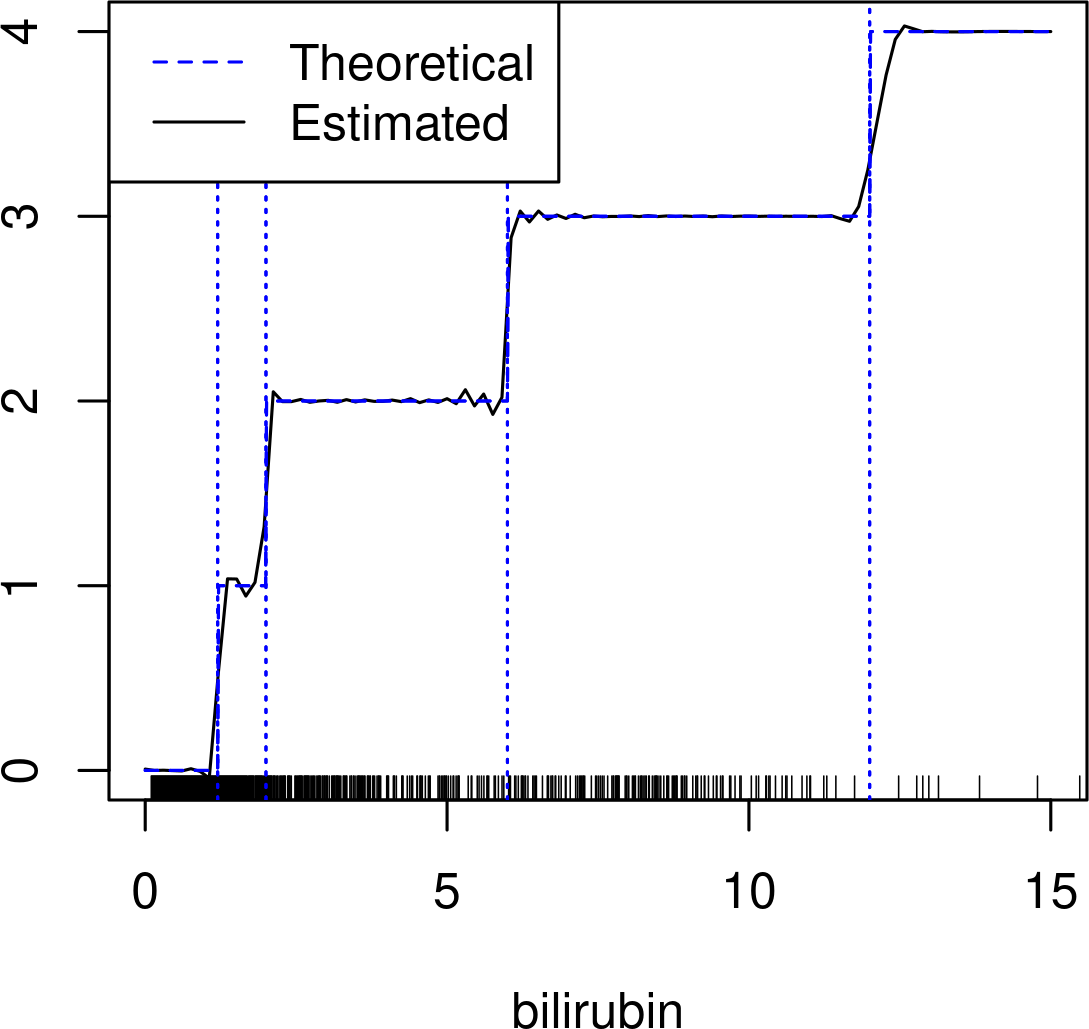}
	\caption{Feature shape of bilirubin feature reconstructing the target labeling function.}
	\label{fig:lsofa_reconstructed_function} 
\end{figure}

Fig \ref{fig:lsofa_reconstructed_function} shows the feature shape of the bilirubin feature for the liver SOFA score for a GAM model with $100$ knots that includes solely the bilirubin feature. Unsurprisingly, the GAM model exactly reconstructs the step function defined by bilirubin intervals. 

\begin{figure}[t!]
	\centering
  		\includegraphics[width=0.7\textwidth]{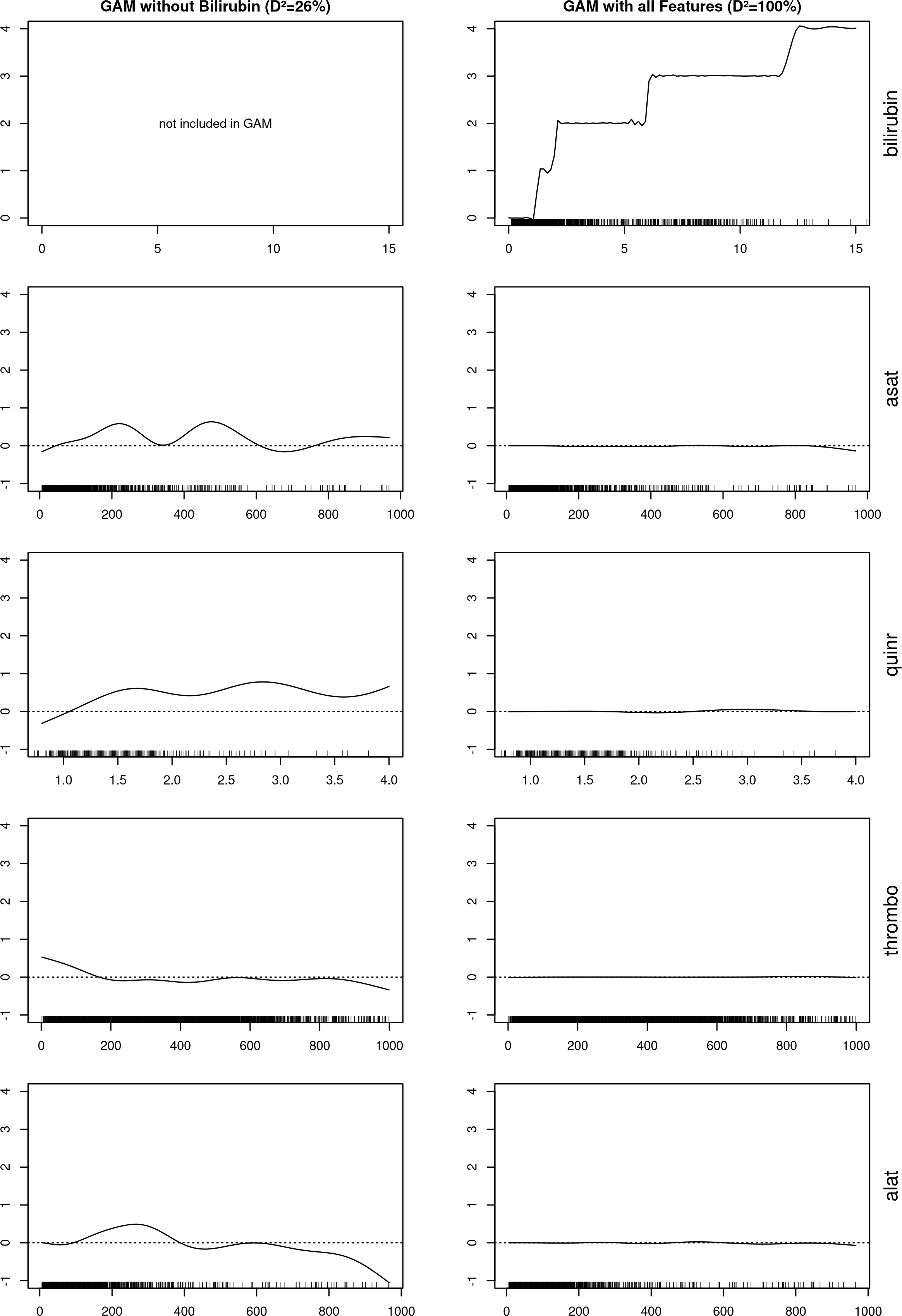}
		\caption{Feature shapes of GAM trained with all features (right column) and without access to bilirubin measurement (left column), showing nullification of non-circular features in the presence of circular features.}
	\label{fig:lsofa_nullification}
\end{figure}

Fig \ref{fig:lsofa_nullification} shows the feature shapes of two more complex GAMs trained on the liver SOFA data. Both models include the features asat, quinr, alat, and hzv, however, with (right column) and without (left column) access to bilirubin measurements. As any model that includes the bilirubin feature, the model in the right column has a $D^2$ values of $100\%$, where the model that includes bilirubin as sole feature has the least degrees of freedom out of all models in the powerset. As shown in the left column of Fig \ref{fig:lsofa_nullification}, a model trained on the four features asat, quinr, alat, and hzv, excluding bilirubin, explains the data at a $D^2$ value of $26\%$ and shows non-neglible contributions of these features. However, as soon as the bilirubin feature is added to the model, the contribution of these features is completely nullified, as seen in the last four rows of the right column. We note that even at an enlarged scale, the feature shapes of the nullified features approximate constant zero lines. This allows us to identify bilirubin as a circular feature in the dataset for liver SOFA score.

A more complex example of the kidney SOFA score is described in Section B1 of the Appendix. The deterministic rule scheme is defined as a step function, depending on the maximum score of two conditions, based on measurements of creatinine and urine output. Despite the increased complexity, circularity of defining features can exactly be identified.

\subsection*{Circularity in machine learning models}

In the case of machine learning prediction we assume that we know the functional definition of the target labels, but we do not have access to the training data $D = \{(\mathbf{x}^n,{y}^n)\}_{n=1}^N$ that were used to optimize the machine learning model. All we have is model predictions on test data $T = \{(\mathbf{x}^m,\hat{y}^m)\}_{m=1}^M$.  The question we would like to answer is whether we can detect, from the test-set predictions of the black-box model alone, whether the model that performed the predictions had access to a feature that allows reconstructing a known deterministic functional relationship. 

\subsubsection*{Circularity in machine learning for patent prior art search.} 

Let us consider an example that is inspired by the KISS principle ("keep it simple and straightforward") applied in patent prior art search \cite{MagdyJones:10}. The idea of this principle is to automatically obtain the citations of a given query patent in a patent retrieval task, and to improve the search by incorporating this information into the search results. \cite{MagdyJones:10} apply this principle in a white-box manner by directly appending IDs of cited patents to the result list of a simpler information retrieval technique such as tf-Idf \cite{Jones:1972}. We are interested in the scenario where the information retrieval model is a black box, i.e, where we can access the machine learning model only via its predictions on test data.  The approach we take is inspired by knowledge distillation where the predictions of a black-box teacher model are used as training data for a GAM student model that is based on all combinations of input features. The circularity test described above is then applied to the GAM with the goal of detecting a circular feature that defines the target label among the candidate input features.

An example for a black-box teacher model is described in \cite{KuwaETAL:20}, who apply a "stacking" approach \cite{Wolpert:92} to combine various model scores into an aggregate similarity score between query and document. In difference to their linear combination, we employ a nonlinear combination of the scores listed in Table \ref{tab:features_clir} using a feedforward neural network (or multi-layer perceptron (MLP)). The feedforward neural network was implemented in \url{pytorch.org}. It consists of $7$ layers, with an ascending, then descending number of neurons per layer, and a tanh activation function. It was trained using PyTorch's SGD optimizer, with batch size $64$, learning rate $.01$, without dropout, for $5$ epochs.  All other optimizer settings are default values of PyTorch's SGD optimizer. In our experiment, the target labels are defined as binary relevance ranks for patent queries, and were constructed deterministically by assigning a relevance level of $1$ for either inventor citation, examiner citation, or family membership, and a relevance level of $0$ for all other documents. 
The parameters of the teacher neural network are trained for logistic regression on $1{,}500$ queries, resulting in $318{,}375$ observations of query-document pairs. The predictions were thresholded at $0.5$ and evaluated with respect to F1 score \cite{ManningETAL:08}. The teacher neural network on the test set of $500$ queries reaches $100\%$ F1. As we will see, this result is too good to be true, since it can be traced back to the teacher neural network reconstructing the deterministic target function while ignoring all other features of the model.

\begin{figure}[t!]
	\centering
	\includegraphics[width=0.8\textwidth]{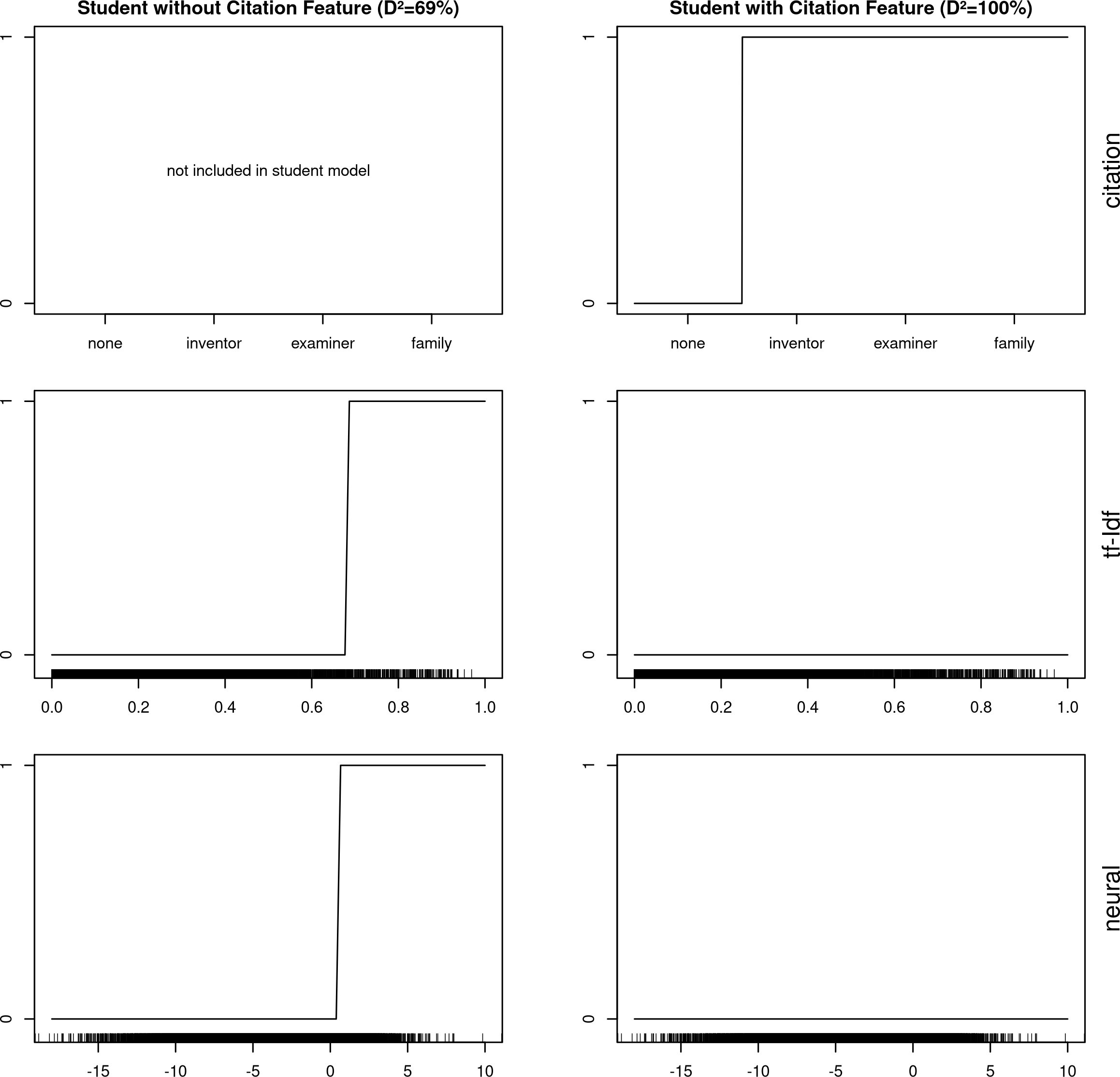}
	\caption{Feature shapes of two student GAMs for the same teacher which had access to all features during training. The student in the right column did have access to citation features, while the student in the left column did not. Features in the presence of citations features are nullified.}
	\label{fig:clir_distill_compare_students}
\end{figure}

As a first circularity check, we fit a student GAM model that has access to all features in Table \ref{tab:features_clir} and treats the neural teacher model's predicted labels similar to gold standard labels. For the binary classification data, we use a GAM that assumes a binomial response variable and a logistic link function. The student model is trained on the $500$ test queries that were annotated with relevance ranks predicted by the teacher neural network, resulting in $106{,}690$ query-document observations. 
Fig \ref{fig:clir_distill_compare_students}, right column, shows a student model including citation features, reaching a $D^2$ value of $100\%$. The three plots in this column show that any model including citation features has learned to rely exclusively on them. A student model that does not include citation features, but only tf-Idf and the neural joint score, is shown in the left column of Fig \ref{fig:clir_distill_compare_students}. It reaches a respectable $D^2$ value of $69\%$ and shows a strong contribution of the tf-Idf and neural joint score features to the prediction. However, the step function feature shapes in the left column are completely nullified in the student model in the right column that includes citation features, shown in the flat lined feature shapes in the right column. This confirms that the teacher model must have incorporated the functional definition of relevance ranks via citations as feature into the model. 

\begin{figure}[t!]
	\centering
	\includegraphics[width=0.8\textwidth]{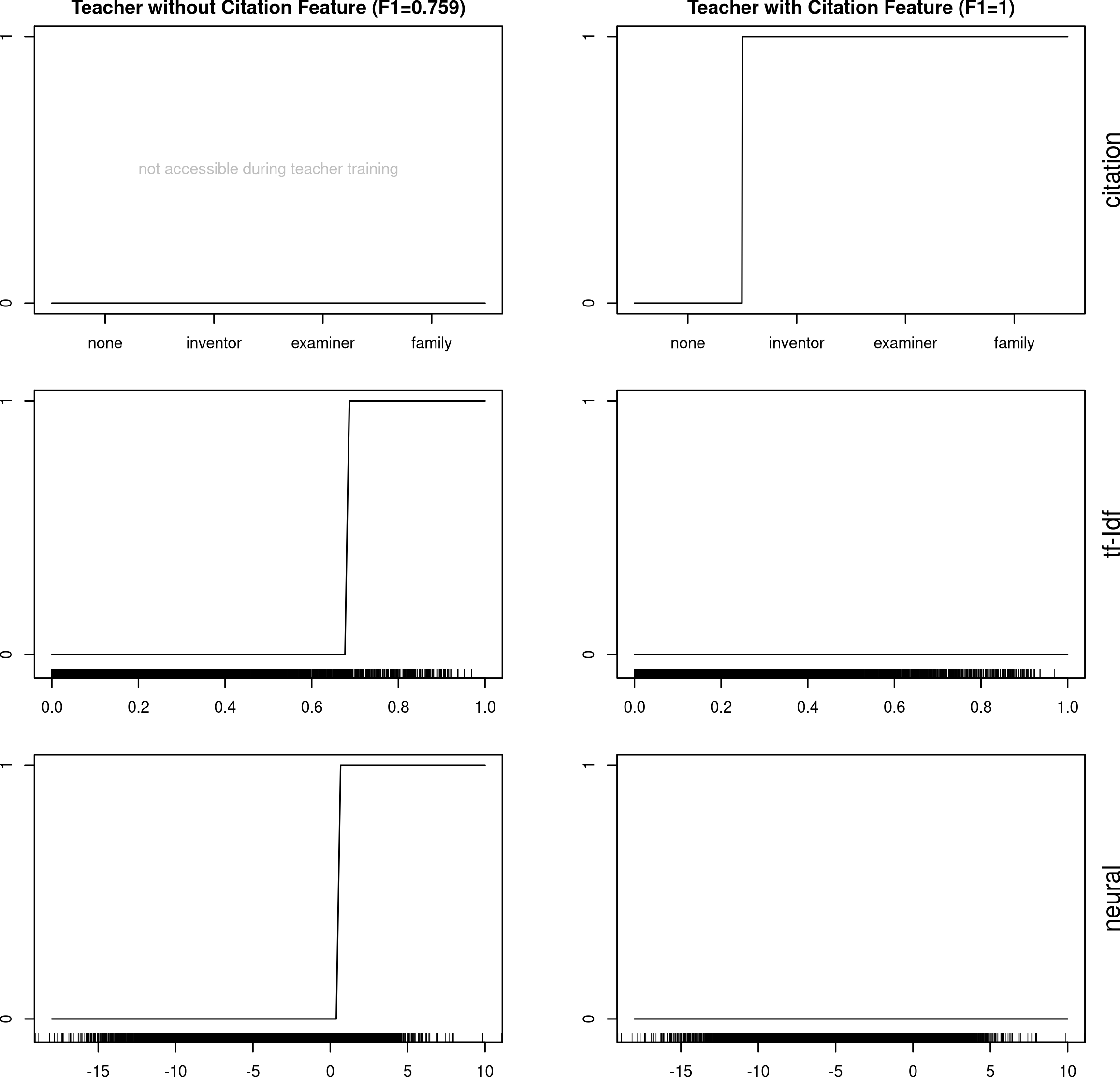}
	\caption{Features shapes of two student GAMs that had access to all features, where the teacher in the right column had access to citation features, while the teacher in the left column did not. Features in the presence of citation features are nullified.}
	\label{fig:clir_distill_compare_teachers}
\end{figure}

In order to confirm that our student GAM is not hallucinating circularity, we conducted a control experiment where we trained a teacher neural network explicitly without access to the citation features. As shown in Fig \ref{fig:clir_distill_compare_teachers}, the teacher without citation information yields an F1 score of $75.9\%$ on the test set, while the teacher with citation information reaches an F1 score of $100\%$. Next we fitted two student GAMs that had access to all features. One GAM was distilled from the teacher trained with citation features, the other GAM was distilled from the teacher trained without citation features. The first GAM is identical to the right column of Fig \ref{fig:clir_distill_compare_students} and repeated in the right column of Fig \ref{fig:clir_distill_compare_teachers}. We can clearly see that the function represented by the teacher with citation features is identical to the deterministic definition of the target.
The student distilled from the teacher without citation access, shown on the left, again confirms a strong contribution of the tf-Idf and neural score features to the prediction, however, this contribution is nullified if the teacher has access to citation features. 

Furthermore, we performed an ablation study where all citation features were set to zero on the test data. This experiment demonstrates the effect of the citation features on the system performance. We observed a dramatic drop in F1 score for the teacher incorporating citation features, from $100\%$ on the test set including citation features to $0\%$ on the ablation test set not including citation information.

\begin{figure}[t!]
 \centering
   \includegraphics[width=0.8\textwidth]{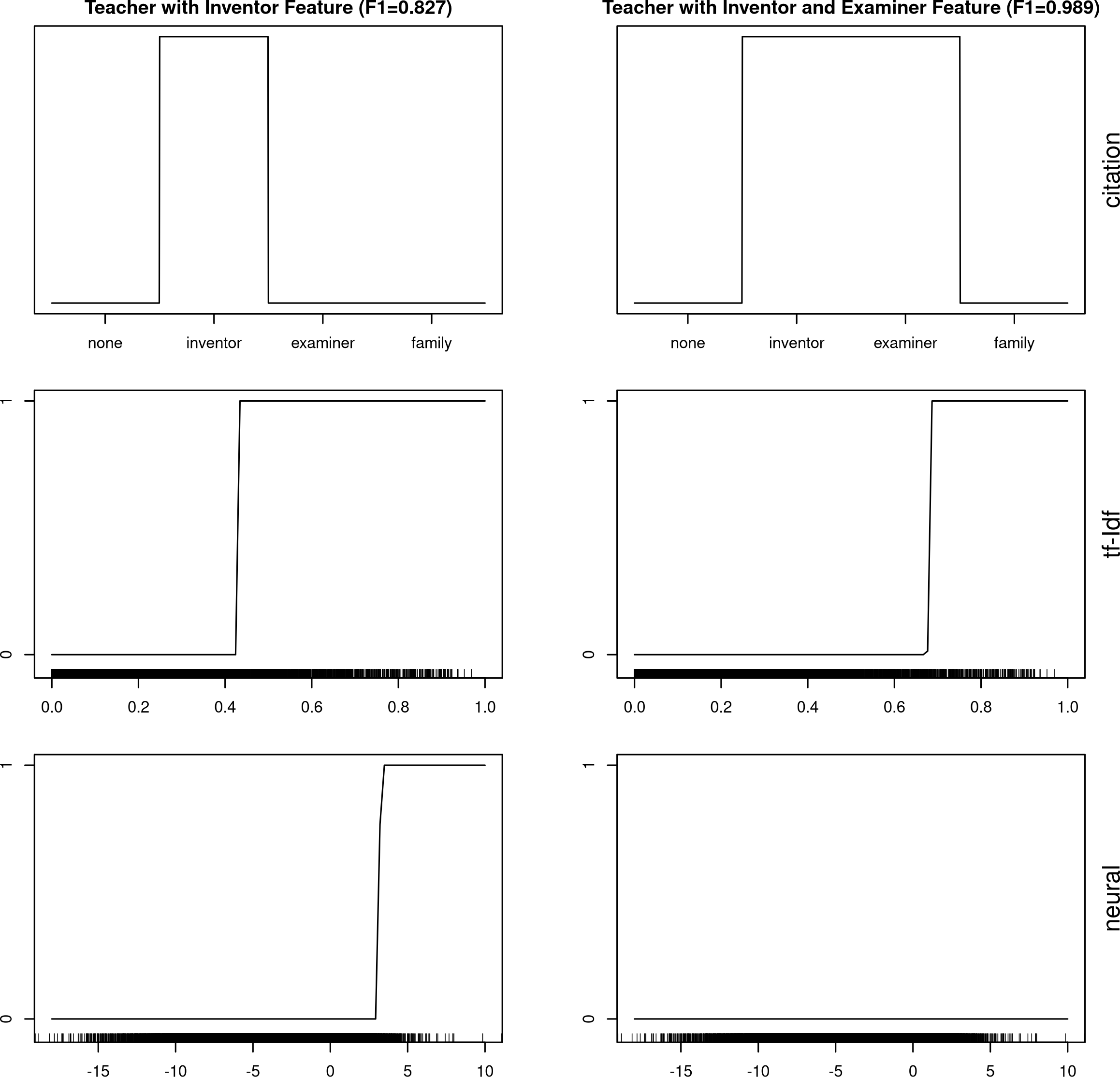}
 \caption{Feature shapes of cross-lingual patent retrieval features and partially circular citation features against relevance scores.}
 \label{fig:distill_part_circularity_clir}
 \end{figure}

An example of \emph{partial circularity} is given in Fig \ref{fig:distill_part_circularity_clir}. In this example, only information about inventor citations (left column) or information about inventor and examiner citations (right column) is included as features for the teacher model. The inclusion of the first type of information is similar to the effect of bias features which are strongly correlated with the target label, but do not suffice to exactly reconstruct the target function. Information about examiner citations can be seen as an illegitimate feature since relevance judgements obtained from patent examiners are "leaks from the future" \cite{KaufmanETAL:11} if patent prior art search is supposed to support the patent examiner. Our analysis shows that a teacher network that has access to inventor and examiner citations reaches an F1-score of nearly $100\%$ on the test set, while a teacher that has access to inventor citations only reaches an F1-score of $83\%$. The feature shapes of the student GAMs trained on predictions of the respective teacher models clearly identify the use of the respective features during training, shown in the top row of Fig \ref{fig:distill_part_circularity_clir}.
For a teacher that uses only inventor citations, the feature shapes of the tf-Idf and neural features of the student GAM still show a strong contribution (second and third row in left column of Fig \ref{fig:distill_part_circularity_clir}). However, a teacher that uses both inventor and examiner citations diminishes the contribution of the tf-Idf feature, shown by a right-shift of the respective feature shape (second row in right column of Fig \ref{fig:distill_part_circularity_clir}) and nullifies the contribution of the neural score feature (third row in right column of Fig \ref{fig:distill_part_circularity_clir}). Since the functional definition of the target relevance labels has only partially been included in training of the teacher model, the features are only partially nullified in the student GAM. 

\subsubsection*{Circularity in machine learning in medical data science.}

\begin{figure}[t!]
\centering
  \includegraphics[width=0.6\textwidth]{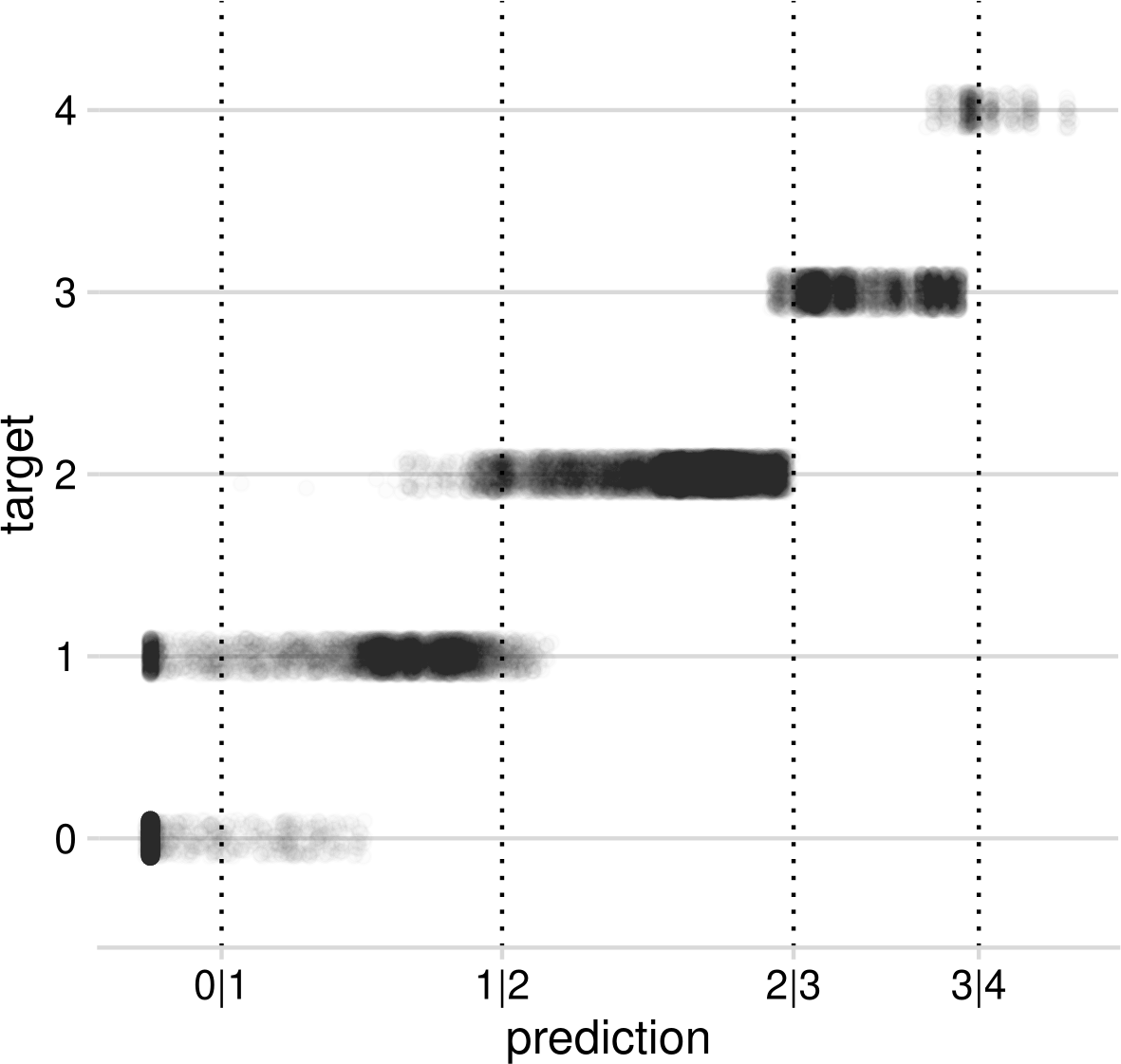}
\caption{Distribution of teacher model score by target class on liver SOFA test set.}
\label{fig:distill_distrib_lsofa}
\end{figure}

\begin{figure}[t!]
\centering
  \includegraphics[width=0.7\textwidth]{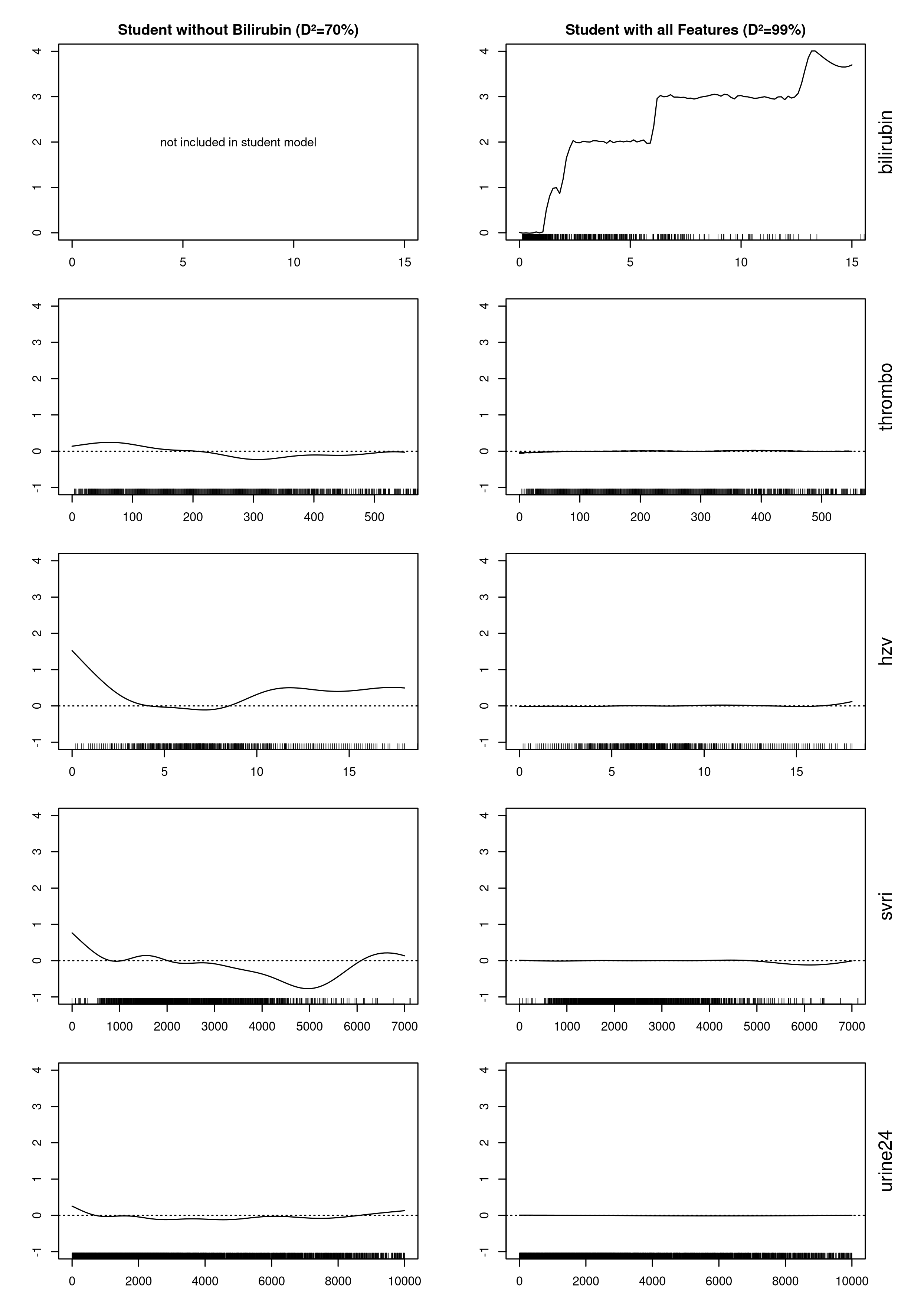}
\caption{Feature shapes of two student GAMs for the same teacher which had access to all features during training. The student in the right column did have access to bilirubin, while the student in the left column did not. Features in the presence of citations features are nullified.}
\label{fig:distill_circularity_lsofa}
\end{figure}

Let us next consider circularity in the prediction of liver SOFA scores. We want know whether we can tell from the test set predictions alone, without knowing the training data, if a neural network is able to reconstruct the functional definition of liver SOFA scores given in Table \ref{tab:definition_lsofa}, and whether the learned predictions of the neural network are only applications of this rule. We again employ knowledge distillation: As a teacher network, we again consider a feedforward neural network. The network was implemented in \url{pytorch.org}. It consists of $7$ layers, with an ascending, then descending number of neurons per layer, and a ReLU activation function \cite{GlorotETAL:11}. It was trained using PyTorch's SGD optimizer, with batch size $64$, learning rate $.01$, and dropout rate of $0.2$ in hidden layers, for $5$ epochs.  All other optimizer settings are default values of PyTorch's SGD optimizer. The nework was trained for regression on $323{,}404$ measurement points of the ICU data for $620$ patients described in \cite{SchamoniETAL:19}. Furthermore, thresholds to turn the real-valued teacher network output scores into discrete SOFA scores were learned. The predictions were tested on another $80{,}671$ measurement points. The train and test data include all $45$ clinical measurements described in \cite{SchamoniETAL:19} as input features and use liver SOFA scores that were assigned automatically following the functional definition in Table \ref{tab:definition_lsofa} as gold standard labels. The accuracy \cite{ManningETAL:08} of the feedforward teacher network on the test data is $98.1\%$, where the most accurate predictions were made for the target scores $0$, $1$, $2$, and $3$, with minor mispredictions for target class $4$ (see Fig \ref{fig:distill_distrib_lsofa}).
Starting from the feature representation of all $45$ clinical measurements described in \cite{SchamoniETAL:19}, we select the five features that are most highly correlated with the label predicted by the teacher model: bilirubin (bili), thrombocytes (thrombo), cardiac output (hzv), systematic vascular resistance index (svri), and urine output (urine). Based on the predictions of the teacher feedforward network and these five features, we train a GAM student model with $100$ knots. As can be seen from Fig \ref{fig:distill_circularity_lsofa}, top right, the objective function of liver SOFA scores can be recreated very accurately by GAM student models including the bilirubin feature with a $D^2$ value of $99\%$. Out of all models, the one that includes only bilirubin as feature has the fewest degrees of freedom. The left column shows that all other features have a strong contribution to the prediction if bilirubin is not included in the model, yielding $D^2$ values of $70\%$. However, as can be seen in the right column, the contribution of all other features except bilirubin is nullified in any model including bilirubin. 
To conclude, we identified the bilirubin feature as sufficient to perform a deterministic prediction of the liver SOFA scores assigned by the teacher neural network, while we ruled out other features as deterministic predictors despite their strong correlation with the predicted target labels. We can therefore assume that the neural network that produced the test set predictions included bilirubin as circular feature during training, and it learned nothing but how to reproduce the known deterministic rule to assign liver SOFA scores based on bilirubin levels.

Similar to circularity in medical data, a more complex example of circularity of machine learning models for the kidney SOFA score is described in Section B.2 of the Appendix. Despite the increased complexity, circularity of defining features can exactly be identified from the predictions of a black-box neural network.

\section*{Discussion}

The examples discussed in this paper use real-world datasets and are based on the building blocks of machine learning algorithms used in benchmark competitions in the fields of cross-lingual patent retrieval \cite{PiroiTait:10} and medical data science \cite{ReynaETAL:19}. We showed that an inclusion of measurements that define target outcomes as features in the input data representations allows machine learning algorithms to reconstruct the known deterministic functional definition of the target, leading to circular predictions that are based solely on what is known beforehand. Our circularity test is a tool for a clear-cut identification of circular features in machine learning data and black-box models such as neural networks. 

Including circular features into a model can happen deliberately or inadvertently. However, in any case they will hinder effectively transferring machine learning expertise to real-world applications in these economically and socially important fields and should clearly be avoided. Firstly, machine learning models trained on data including the defining measurements for target outputs will yield nearly perfect predictions on input data including the defining measurements, but they cannot be transferred to unseen data where the defining features are not or only incompletely available. Secondly, a circular learning setup that nullifies the contribution of all features except those defining the target dashes the hope to detect features that could shed new light on predictive patterns. 
In order to avoid validity problems by circular features, any dataset provider or organizer of a benchmark testing challenge should explicitly disclose the functional definition of target labels if a deterministic rule was used in the dataset creation. Furthermore, participants in the benchmark testing challenge should be warned not to include features that deterministically define the target labels in their models. Another solution is to rely on implicit expert knowledge in data annotation instead of following automatic data annotation via deterministic functional definitions. Such an approach has been presented by \cite{SchamoniETAL:19} for the area of sepsis prediction. Here the gold standard labels are obtained in the form of an electronic questionnaire which records attending physicians' daily judgements of patients' sepsis status, thus exploiting implicit knowledge of clinical practitioners. As shown in \cite{SchamoniETAL:19}, the $\kappa$ agreement coefficient between expert labels and algorithmically generated Sepsis-3 labels is $0.34$, which is to be considered minimal or weak agreement. This shows that even if one could argue that expert decisions are potentially influenced by known sepsis definitions, the circularity issue in this setup is minimal. Machine learning based on such non-circular data then allows detecting potentially surprising findings, such as the identification of increased sepsis risk with higher concentration levels of thrombocytes, contradicting the SOFA-based Sepsis-3 definition, but in accordance with other research on sepsis \cite{StoppelaarETAL:14}. 

What are further potential cases of circularity beyond those discussed here, or in other words, how likely is it that other datasets and machine learning models exhibit a yet undetected circularity problem? Critical candidates are machine learning applications in empirical sciences like medicine that define the objects of their research, e.g., diseases, by rigid measurement procedures. We conjecture that any disease prediction task in medical data science needs to be extra cautious to keep measurements that define target outcomes separate from data representations for machine learning. The same is true for any other prediction task based on data derived from measurements. 

In sum, we conjecture that promising results for machine learning prediction may well be achieved even in circular setups. However, cautious analyses are required in order to discern whether the machine learning prediction is based on an accurate estimate of the known functional definition from data --- something which could have been achieved easier by a programmatic application of the deterministic thresholding function --- or whether the actual strength of machine learning --- generalizing predictions based on patterns that go beyond known deterministic rules --- has been exploited.

\section*{Acknowledgments}

This research has been conducted in project SCIDATOS (Scientific Computing for Improved Detection and Therapy of Sepsis), funded by the Klaus Tschira Foundation, Germany (Grant number 00.0277.2015).



\newpage

\appendix

\section*{Appendix}

\section{Mathematical Background}

\subsection{Nullification}

The \emph{nullification} criterion used in the circularity test makes crucial use of the consistency property of the maximum likelihood estimator used to fit GAMs (see \cite{Wood:17}).
\begin{definition}[Consistency]
 Let $M \colon= \left\{ p_\theta \colon \theta \in \Theta\right\}$ be a parametric statistical model where $\theta \mapsto p_\theta$ is injective. Further let $p_{\theta_0} \in M$ denote the true model of the data generating process for a dataset $D = \{(\mathbf{x}^n,{y}^n)\}_{n=1}^N$. Then an estimator ${\theta}_N$ is called \textit{consistent} iff for all $\epsilon > 0$ holds 
 	\begin{align*}
 		P\left( | {\theta}_N - \theta_0 |  > \epsilon \right)  \xrightarrow[N \to \infty]{} 0.
	\end{align*}
\end{definition}

An important consequence of the consistency property of GAMs is that it allows us to differentiate features that deterministically define target labels from irrelevant features by the additive structure of feature shapes represented by the trained GAM. 
\begin{proposition}[Nullification]
\label{prop:nullification}
Let $p^{\textrm{GAM}}_{{\theta}_N}$ be a GAM that optimizes the likelihood of data $D = \{(\mathbf{x}^n,{y}^n)\}_{n=1}^N$ that have been produced by a deterministic data labeling function $p: \mathbf{x}^n \to y^n, \; n=1, \ldots, N$. Furthermore, assume that $p$ can be approximated by an identifiable statistical model $M^{\textrm{GAM}} = \{ p^{\textrm{GAM}}_{\theta}: \theta \in \Theta \}$. Then any feature $x_k$ determining the data generation process will have a non-zero feature shape $f(x_k)$ approximating the the deterministic labeling function $p(x_k)$, and any additional features $x_j, j \neq k$ will have a feature shape of a constant zero function, with a probability that converges to $1$ as the sample size increases. 
\end{proposition}

\textbf{\emph{Proof sketch.}} The proposition follows directly from the consistency of maximum likelihood estimators for GAMs. This has been shown, for example, by \cite{Heckman:86} for GAMs based on cubic regression splines. By consistency, the maximum likelihood estimator $\theta_N$ will converge in probability to the data generating parameters $\theta_0$. Since the model $M^{\textrm{GAM}} = \{ p^{\textrm{GAM}}_{\theta}: \theta \in \Theta \}$ is identifiable, by the injectivity of the mapping $\theta \mapsto p_\theta$, the data generating parameters $\theta_0$ will identify the data generating model $p^{\textrm{GAM}}_{{\theta}_0}$. In this model, only features determining the feature-label relations in the data $D = \{(\mathbf{x}^n,{y}^n)\}_{n=1}^N$ have non-zero feature shapes, and the feature shapes of all additional features have constant zero values. $\square$

\section{Further Experiments}

\subsection{Circularity in medical data for kidney SOFA score}

Identification of circularity of features in data for kidney SOFA score requires recovering a deterministic rule scheme that is defined as a step function of the kidney status, depending on the maximum score of two conditions, based on measurements of creatinine and urine output (see Table \ref{tab:definition_sofa}). The features in the following experiment consist of clinical measurements of creatinine (crea), urine output in the previous $24$ hours (urine24), pH-value of the arterial blood (artph), blood urea nitrogen (bun), body temperature (temp) and serum lactate (lactate).

\begin{table}[t!]
\caption{Definition of kidney SOFA score based on creatinine and urine levels.}
\label{tab:definition_sofa}
\begin{center}
\begin{tabular}{ccc}
  \toprule
  Condition 1 & Condition 2 & kidney SOFA Score \\
  \midrule
  $0 < $ creatinine $\leq 1.2$ & $500 < $ urine & $0$\\
  $1.2 < $ creatinine $\leq 1.9$ &   & $1$\\
  $1.9 < $ creatinine $\leq 3.4$ &  & $2$\\
  $3.4 < $ creatinine $\leq 4.9$ & $200 < $ urine $\leq 500$ & $3$\\
  creatinine $> 4.9$ & $0 < $ urine $\leq 200$ & $4$\\
  \bottomrule
  \end{tabular}
\end{center}
\end{table}

\begin{figure}[t!]
\centering
  \includegraphics[width=0.5\textwidth]{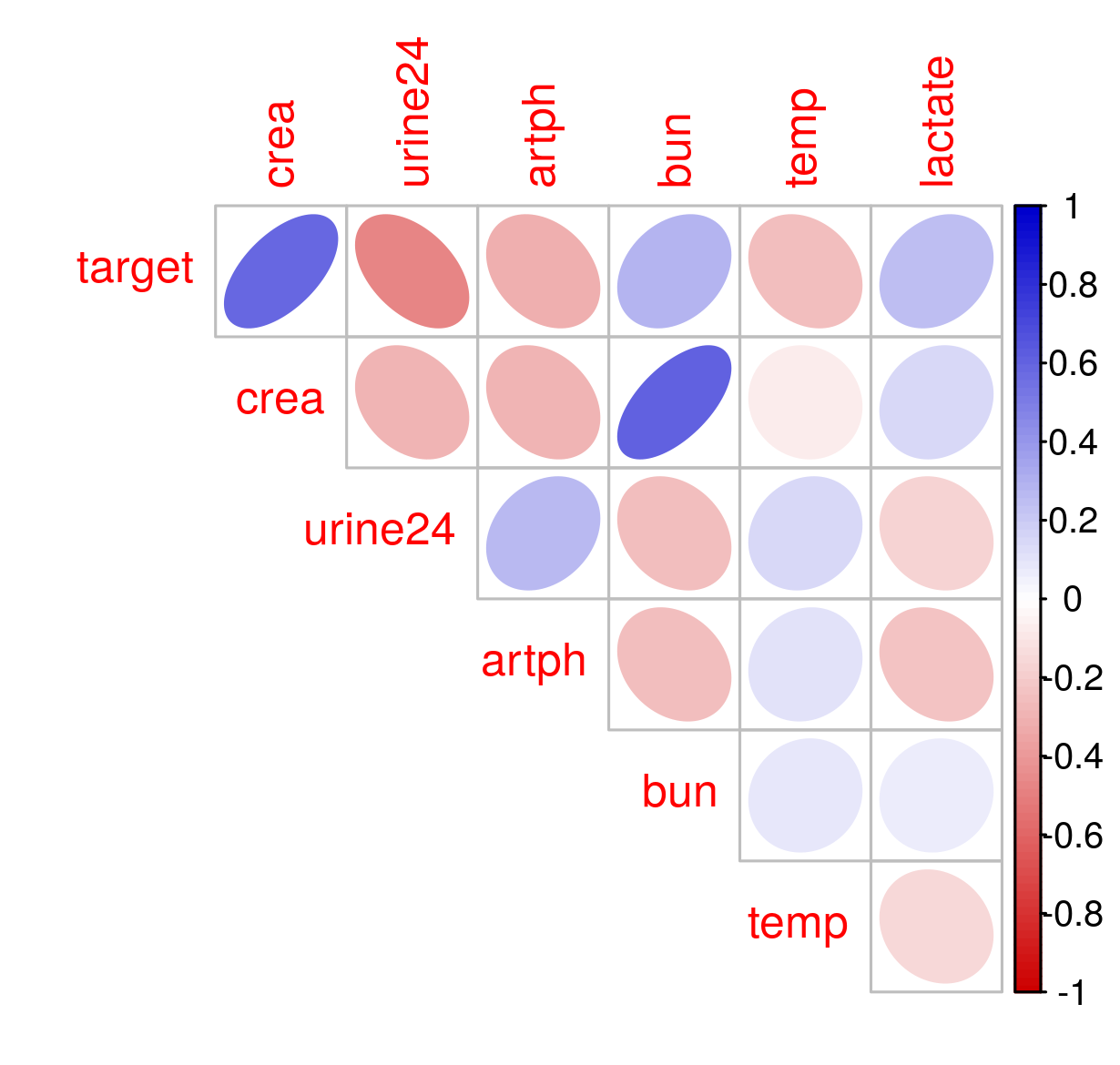}
\caption{Correlation analysis of clinical measurements with kidney SOFA score.}
\label{fig:correlation_sofa} 
\end{figure}

\begin{figure}[t!]
\centering
  \includegraphics[width=0.9\textwidth]{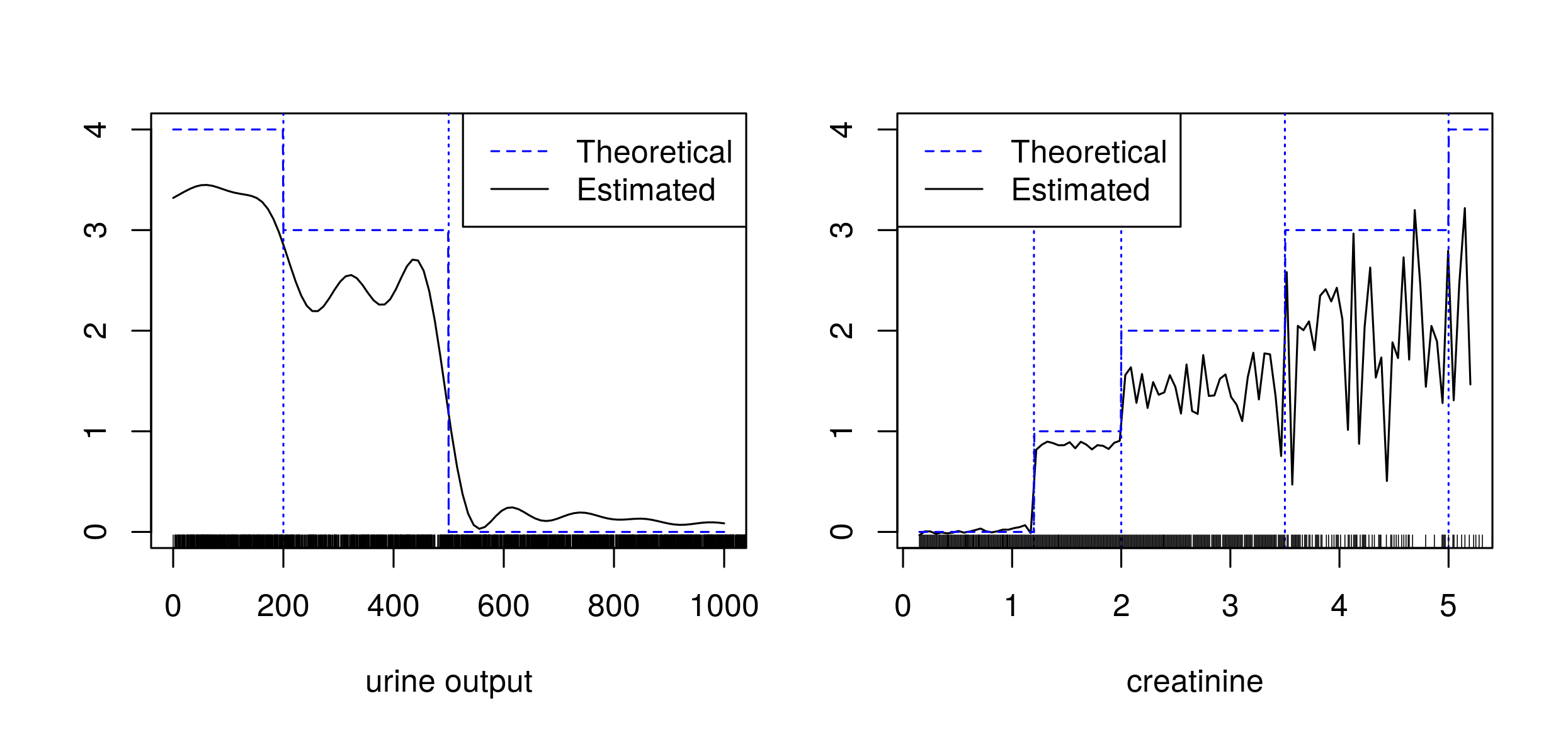}
\caption{Feature shape of creatinine and urine features reconstructing the target labeling function.}
\label{fig:shape_sofa} 
\end{figure}

The correlation analysis shown in Fig \ref{fig:correlation_sofa} demonstrates that all of the above defined features are moderately correlated with kidney status.
Thus a distinction between ``circular'' and simply ``strongly correlated'' features requires thresholding and is an improper tool to assess invalidity. 

\begin{figure}[t!]
\centering
  \includegraphics[width=0.5\textwidth]{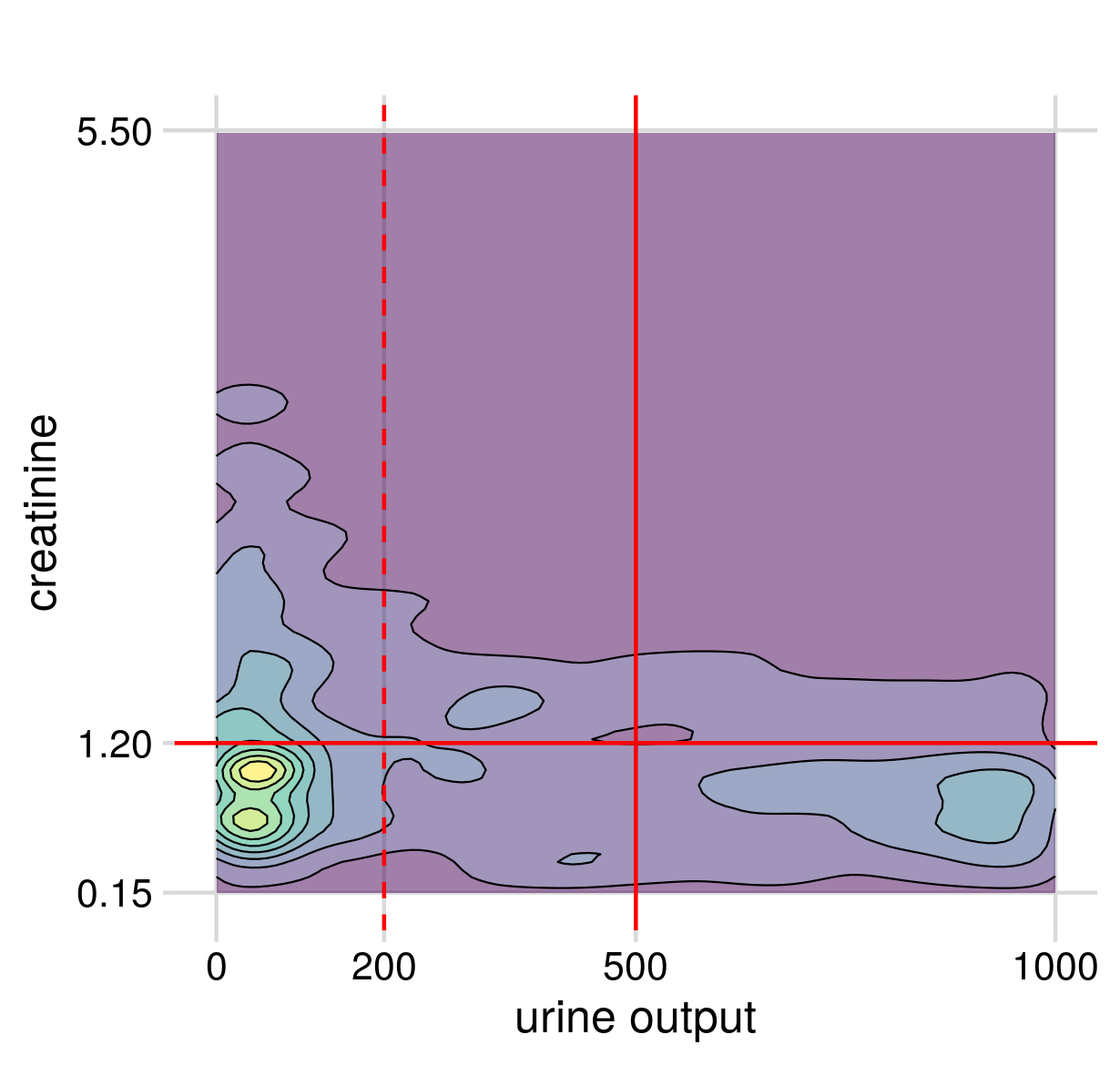}
\caption{Bivariate distribution of urine and creatinine output in data.}
\label{fig:bivariate_sofa} 
\end{figure}

\begin{figure}[t!]
\centering
  \includegraphics[width=0.75\textwidth]{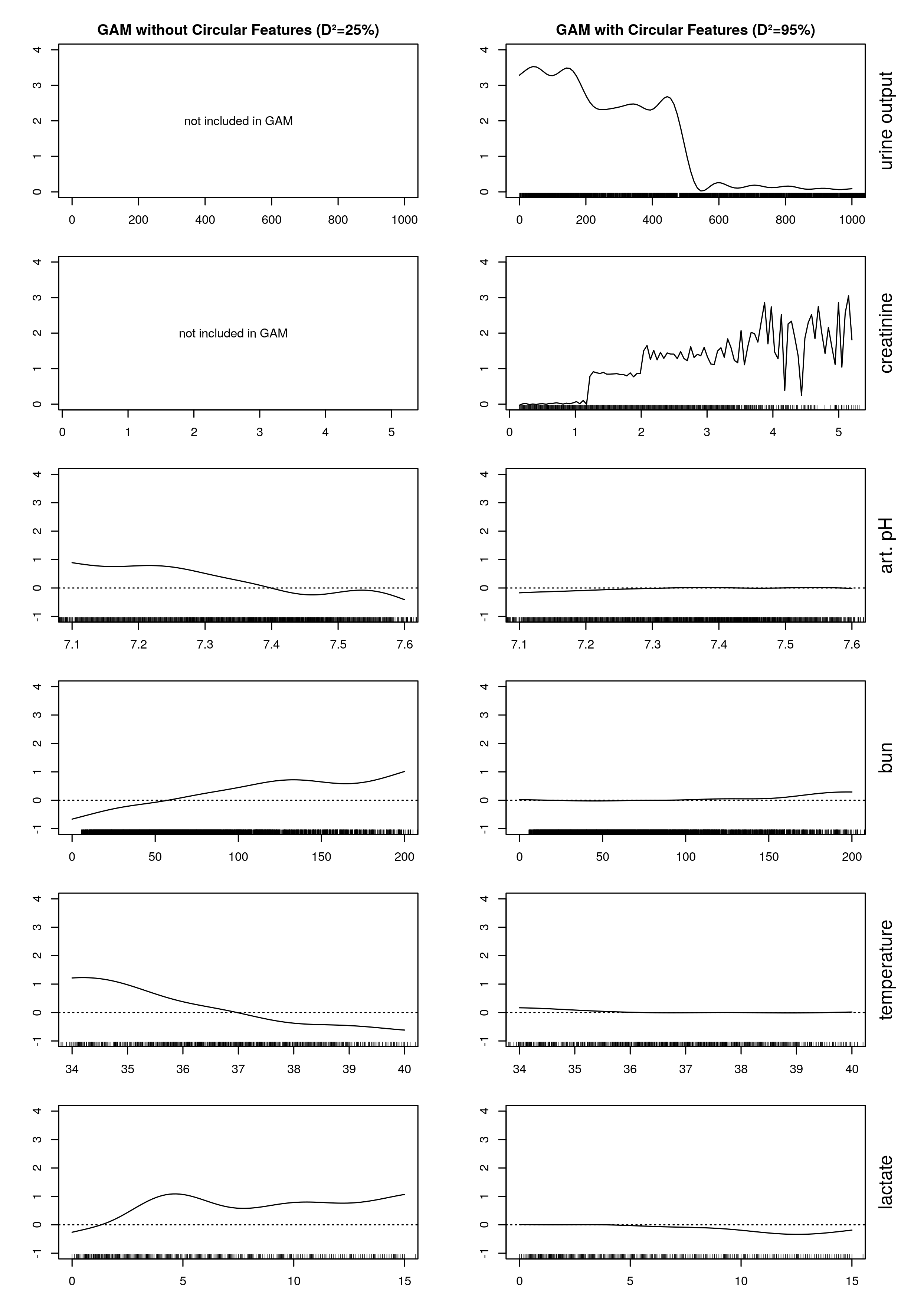}
\caption{Feature shapes of GAM trained with all features (right column) and without access to urine and creatinine measurements (left column), showing nullification of non-circular features in the presence of circular features.
}
\label{fig:circularity-sofa}
\end{figure}

Training a GAM consisting of the single feature of urine output on the ICU data shows that the feature shape of the urine feature almost exactly models the theoretical step function (Fig \ref{fig:shape_sofa}, left plot). The GAM trained with creatinine as the only feature (Fig \ref{fig:shape_sofa}, right plot) shows a less perfect fit of the theoretical step function. This is indicated by the empty tassels in the rug plot for the creatinine feature shape. As can be seen by inspecting the bivariate distribution of creatinine and urine in our data in Fig \ref{fig:bivariate_sofa}, most data points with critically high creatinine level also have a critical low urine level, thus the variables are highly confounded. The reason for the suboptimal fit of the theoretical step function in the case of creatinine is data sparsity in the areas of urine output $> 500$ and creatinine $> 1.2$. These are the data area where a high kidney status would be caused solely by high creatinine levels. However, there are enough data points across the whole range of urine outputs so that a satisfactory fit of the theoretical step function is possible by the feature shape of the urine feature. Still, the highest $D^2$ value of $95\%$ with the fewest degrees of freedom out of all models is obtained by a model including only creatinine and urine as features, thus serving as a strong indicator for circularity of this feature set.

Fig \ref{fig:circularity-sofa} compares the feature shapes of a model using all six features (right column) with the features shapes of a model that excludes the candidate circular features of creatinine and urine (left column). The top two plots in the right column are identical to the feature shapes of urine and creatinine shown in Fig \ref{fig:shape_sofa}. The $D^2$ value of the models on the right reach $95\%$, compared to $25\%$ for the models without circular features. The bottom four plots show that the contribution of any feature in the model without creatinine and urine (left column) is nullified by inclusion of urine and creatinine as features (right column). We note that even at an enlargened scale, the feature shapes of the nullified features approximate constant zero lines. This again confirms the identification of urine and creatinine as circular features in the dataset for SOFA score. 

\subsection{Circularity in machine learning predictions of kidney SOFA score}

\begin{figure}[t!]
\centering
  \includegraphics[width=0.4\textwidth]{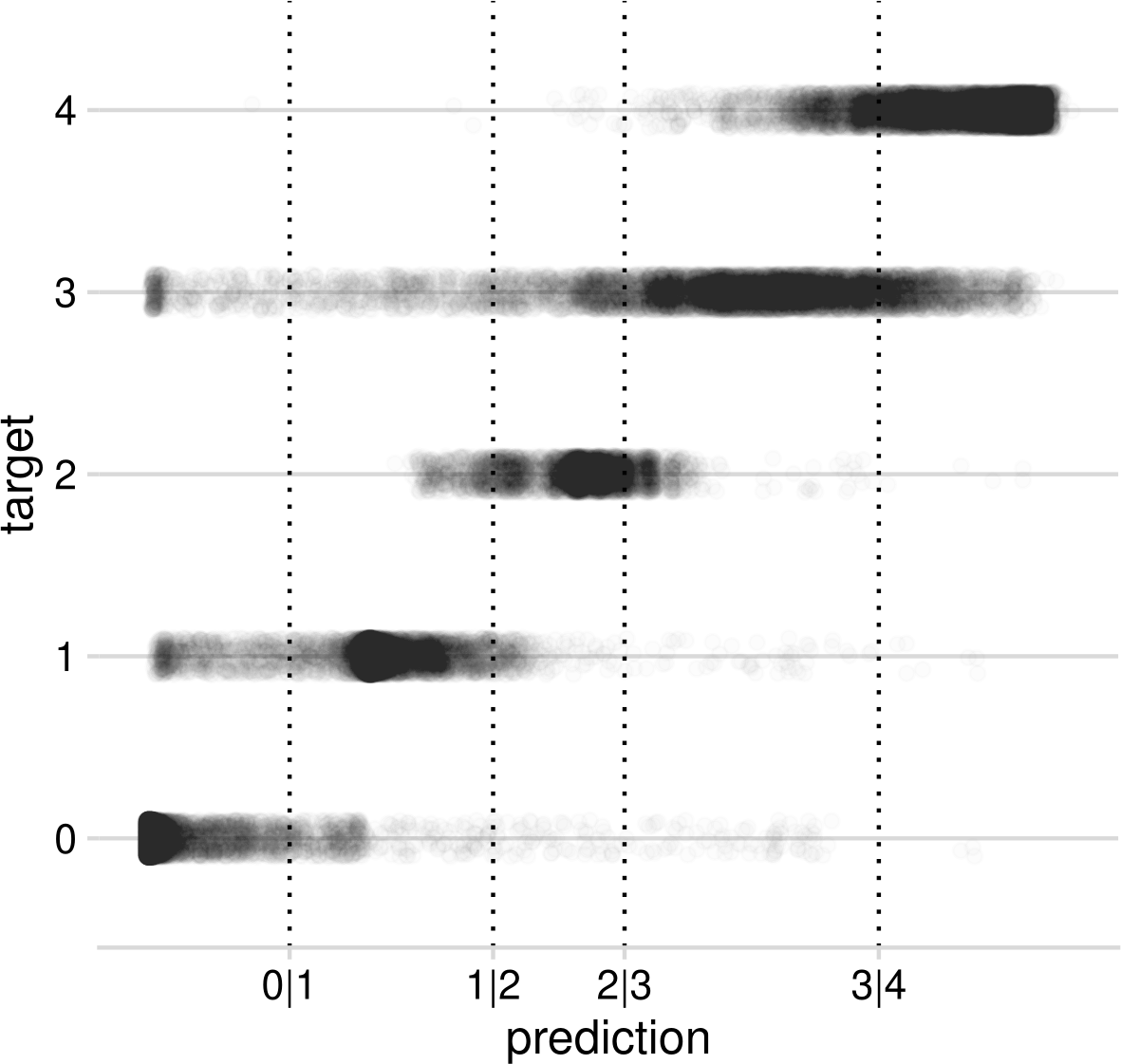}
\caption{Distribution of teacher model score by target class on kidney SOFA test set.}
\label{fig:distill_distrib_sofa}
\end{figure}

\begin{figure}[t!]
\centering
  \includegraphics[width=0.7\textwidth]{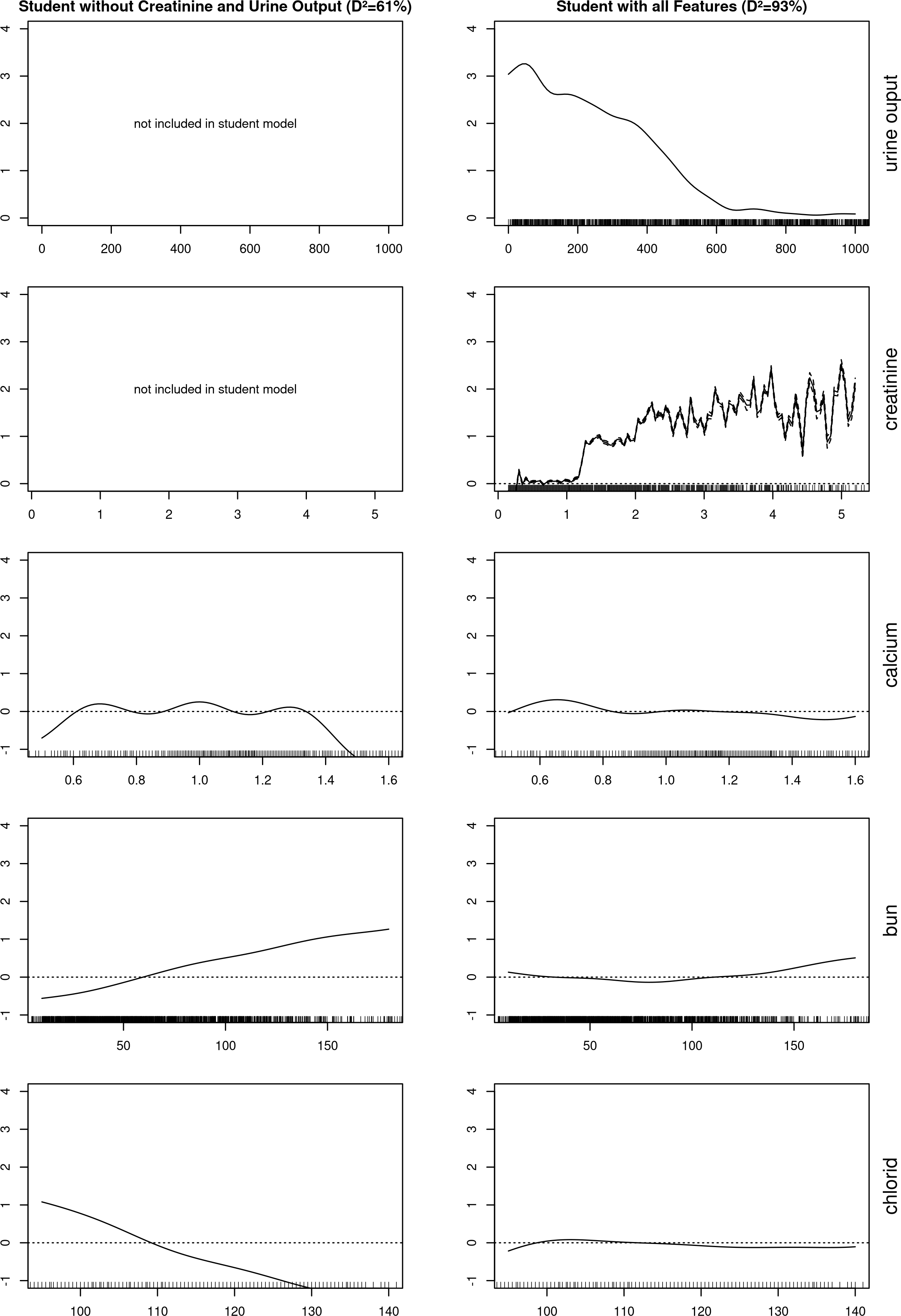}
\caption{Feature shapes of two student GAMs for the same teacher which had access to all features during training. The student in the right column did have access to creatinine and urine features, while the student in the left column did not. Features in the presence of citations features are nullified.
}
\label{fig:distill_circularity_sofa}
\end{figure}

Let us further consider circularity in model prediction for the more complex kidney SOFA score. As a teacher model, we train the feedforward neural network described above\footnote{Minor differences in hyper-parameter settings to the model trained for liver SOFA prediction include a smaller batch size of $32$ and a dropout rate of $0$.} for regression on automatically assigned kidney SOFA labels, following the functional definition in Table \ref{tab:definition_sofa}. The accuracy of the teacher feedforward network on the test labels is $92.2\%$, with minor misclassifications happening for target scores $2$ and $3$ (see Fig \ref{fig:distill_distrib_sofa}).
Out of the $45$ input features, we select the five features that are most highly correlated with the kidney SOFA score predicted by the teacher model. 
These are the clinical measurements of creatinine (crea), urine output in the previous $24$ hours (urine24), pH-values of arterial blood (artph), blood urea nitrogen (bun), and bilirubin (bili).
We train a student GAM with up to $230$ knots on the these five features and the labels predicted by the teacher model. 
As can be seen in the top two images on the right of Fig \ref{fig:distill_circularity_sofa}, the theoretical step function can be reasonably approximated in the areas where enough training data points are available. For example, a model trained on the single feature urine output (top right plot in Fig \ref{fig:distill_circularity_sofa}) appropriately represents the theoretical steps by the function estimate. A sketchier approximation is obtained from a model with creatinine as the only feature (second plot on the right of Fig \ref{fig:distill_circularity_sofa}). This is due to data sparsity, indicated by the empty tassels in the rug plot. However, the model with the fewest degrees of freedom out of all models and highest $D^2$ value of $93\%$ is the one that only includes urine and bilirubin as features (right column), whereas models without creatinine and urine (left column) reach a $D^2$ value of $61\%$. Furthermore, the contributions of features like bilirubin, bun, or artph, as shown by the feature shapes in the left column, are nullified if creatinine and urine are included in the model, as shown in the third to fifth image in the right column of Fig \ref{fig:distill_circularity_sofa}. 
Again, this confirms that creatinine and urine are circular features that can deterministically predict the target labels assigned by the teacher neural network. We can therefore assume that the teacher neural network must have included creatinine and urine as features during training, and it thus learned nothing besides how to reproduce the known deterministic definition of kidney SOFA scores based on thresholds of creatinine and urine.

\end{document}